\documentclass[final]{cvpr}

\usepackage{times}
\usepackage{epsfig}
\usepackage{graphicx}
\usepackage{amsmath}
\usepackage{amssymb}

\usepackage[normalem]{ulem}
\usepackage{amsfonts}
\usepackage{multirow}
\usepackage[switch]{lineno}
\usepackage[dvipsnames]{xcolor}
\usepackage{enumitem}
\definecolor{ForestGreen}{RGB}{34,139,34}
\definecolor{oceanboatblue}{rgb}{0.0, 0.47, 0.75}

\usepackage[pagebackref=true,breaklinks=true,colorlinks,bookmarks=false]{hyperref}

\def\cvprPaperID{2590} 
\def\confYear{CVPR 2021}

\begin{document}

\title{Scene-Intuitive Agent for Remote Embodied Visual Grounding}


%
%

\author{Xiangru Lin$^{1}$\quad\quad Guanbin Li$^{2}$\thanks{Corresponding author is Guanbin Li. This work was partially supported by National Key Research and Development Program of China (No.2020YFC2003902). This work was supported in part by the Guangdong Basic and Applied Basic Research Foundation under Grant No.2020B1515020048, in part by the National Natural Science Foundation of China under Grant No.61976250 and No.U1811463. This work was also sponsored by CCF-Tencent Open Research Fund.} \quad\quad \quad\quad Yizhou Yu$^{1,3}$ \vspace{2mm}\\
$^1$The University of Hong Kong  \quad\quad\quad $^2$Sun Yat-sen University \quad\quad\quad $^3$Deepwise AI Lab\\
{\tt\footnotesize xrlin2@cs.hku.hk}, {\tt\small liguanbin@mail.sysu.edu.cn}, {\tt\small yizhouy@acm.org}
	\vspace{-4mm}
}


\maketitle

\begin{abstract}
Humans learn from life events to form intuitions towards the understanding of visual environments and languages. Envision that you are instructed by a high-level instruction, \emph{``Go to the bathroom in the master bedroom and replace the blue towel on the left wall''}, what would you possibly do to carry out the task? Intuitively, we comprehend the semantics of the instruction to form an overview of \textbf{\emph{where}} a bathroom is and \textbf{\emph{what}} a blue towel is in mind; then, we navigate to the target location by consistently matching the bathroom appearance in mind with the current scene. In this paper, we present an agent that mimics such human behaviors. Specifically, we focus on the Remote Embodied Visual Referring Expression in Real Indoor Environments task, called REVERIE, where an agent is asked to correctly localize a remote target object specified by a concise high-level natural language instruction, and propose a two-stage training pipeline. In the first stage, we pre-train the agent with two cross-modal alignment sub-tasks, namely the Scene Grounding task and the Object Grounding task. The agent learns \textbf{\emph{where}} to stop in the Scene Grounding task and \textbf{\emph{what}} to attend to in the Object Grounding task respectively. Then, to generate action sequences, we propose a memory-augmented attentive action decoder to smoothly fuse the pre-trained vision and language representations with the agent's past memory experiences. Without bells and whistles, experimental results show that our method outperforms previous state-of-the-art(SOTA) significantly, demonstrating the effectiveness of our method.
\end{abstract}

\section{Introduction}
\noindent Vision and Language tasks, such as Vision-and-Language Navigation (VLN)~\cite{mattersim}, Visual Question Answering (VQA)~\cite{VQA,cao2018visual} and Referring Expression Comprehension (REF)~\cite{KazemzadehOrdonezMattenBergEMNLP14,yang2020graph,yang2020relationship} etc., have been extensively studied in the wave of deep neural networks. In particular, VLN~\cite{mattersim, chen2019touchdown} is a challenging task that combines both natural language understanding and visual navigation. Recent works have shown promising performance and progress. They mainly focus on designing agents capable of grounding fine-grained natural language instructions, where detailed information is provided, to find \textbf{\emph{where}} to stop, for example \emph{``Leave the bedroom and take a left. Take a left down the hallway and walk straight into the bathroom at the end of the hall. Stop in front of the sink''} ~\cite{fried2018speaker, ma2019selfmonitoring, wang2018look, wang2019reinforced, tan2019envdrop, ke2019tactile}. However, a practical issue is that fine-grained natural language instructions are not always available in real life and human-machine interactions are mostly based on high-level instructions such as \emph{``Go to the bathroom at the end of the hallway''}. In other words, designing an agent that could perform high-level natural language interpretation and infer the probable target location using knowledge of the environments is of more practical use.

In this paper, we focus on the REVERIE task~\cite{qi2020reverie} which is an example of the above mentioned high-level instruction task. Here, we briefly introduce the settings. Given a high-level instruction that refers to a remote target object at a target location within a building, a robot agent spawns at a starting location in the same building and tries to navigate closer to the object. The output of the task is a bounding box encompassing the target object. The success of the task is evaluated based on explicit object grounding at the correct target location. A straightforward solution is to integrate SOTA navigation model with SOTA object grounding model. This strategy has proven to be inefficient in ~\cite{qi2020reverie} and instead, they proposed an interactive module to enable the navigation model to work together with the object grounding model. Although the performance is improved, we observe that such method has a key weakness: it is unreasonable to discern high-level instruction by directly borrowing the fine-grained instruction navigation model that consists of simple trainable language attention mechanism based on the fact that the perception of high-level instruction primarily depends on commonsense knowledge prior as well as past experiences in memory. Therefore, the overall design is not in line with human intuitions in high-level instruction navigation.

\begin{figure}[ht]
  \centering
  \includegraphics[width=\linewidth]{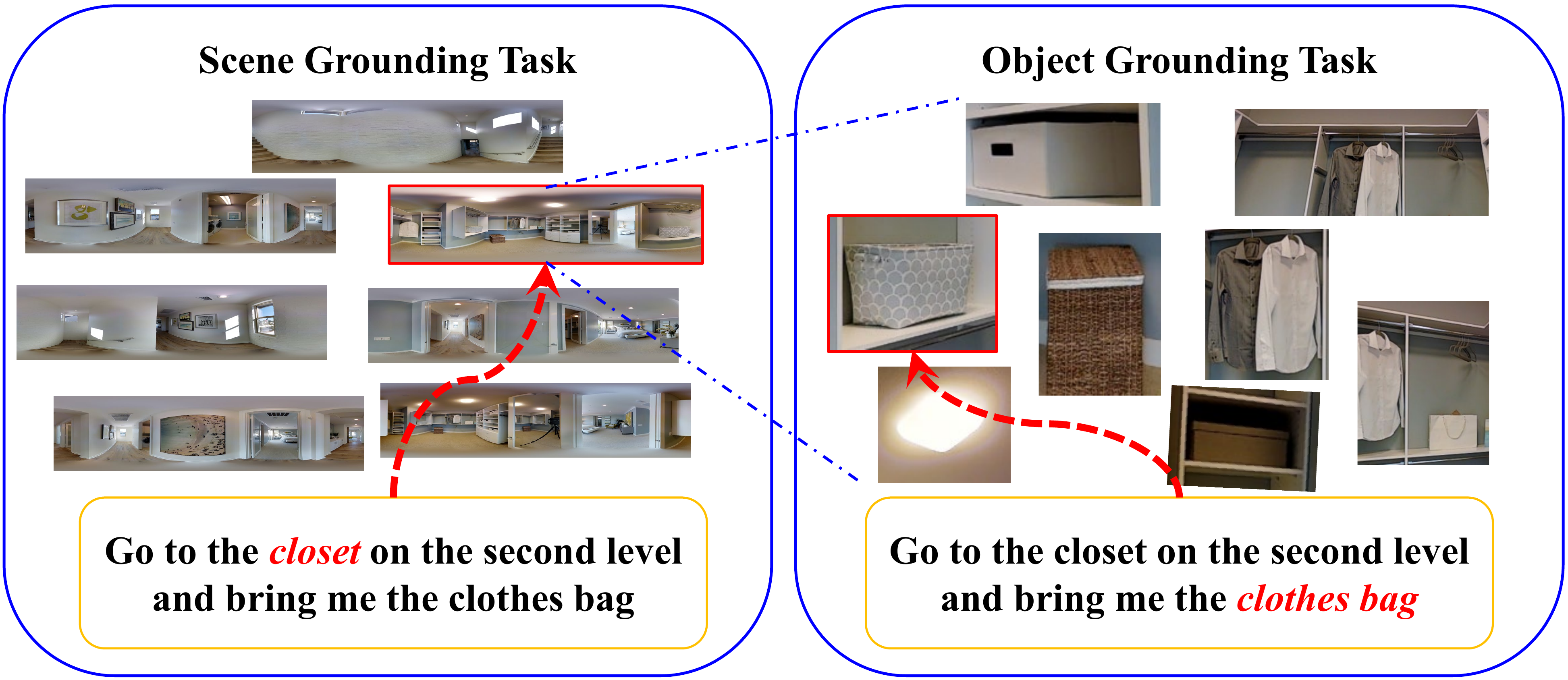}
  \caption{The overview of two pre-training tasks, the Scene Grounding task and the Object Grounding task. The Scene Grounding task empowers the agent the ability to reason where the target location is and the Object Grounding task learns what to attend to.}
  \label{Fig:pretraining}
\end{figure}

Designing an agent to solve the problem like the REVERIE task is still under explored and there are still no systematic ways to design such an agent. Then, how does human wisdom solve this task? Human beings have instincts to understand surrounding visual environments and languages. Intuitively, given a high-level instruction, we would first extract high-level \textbf{\emph{what}} and \textbf{\emph{where}} information and then form an overview of the appearance of the target location in mind based on common sense knowledge. During navigation, we would consistently match current scene and objects in the scene to the instruction semantics and decide where to navigate next. According to such intuitions, we approach this problem from a new perspective and present an agent that imitates such human behaviors. Concretely, we define our problem as designing an agent that is able to solve \textbf{\emph{where}} and \textbf{\emph{what}} problem in the REVERIE task. We propose a two-stage training pipeline. In the first stage, we design two pre-training tasks, mimicking the aforementioned two human intuitions. The second stage is training the agent with a memory-augmented attentive action decoder, further increasing the agent's navigation capability under high-level instructions.

\textbf{Pre-training Stage.} As is shown in Fig.~\ref{Fig:pretraining}, we introduce a new subtask called the Scene Grounding task that is trained to recognize which viewpoint in a set of viewpoints is best aligned with the high-level instruction and another subtask called the Object Grounding task that helps the agent identify the best object that matches to the instruction among a set of candidate objects located at a target viewpoint. Experimental results show that the Scene Grounding model recognizes the target viewpoint with a high accuracy and the Object Grounding model outperforms the previous best model used in ~\cite{yu2018mattnet, qi2020reverie} by more than $10\%$.

\textbf{Action Decoding Stage.} In this stage, with the pre-trained models serving as scene and language encoders, we propose a memory-augmented attentive action decoder that leverages a scene memory structure as the agent's internal past state memory. This design is based on the fact that the computation of action at a specific time step could depend on any provided information in the past. Experimental results indicate that the proposed structure is effective and achieves new state-of-the-art performance.

To sum up, this paper has the following contributions:
\begin{itemize}[noitemsep, nolistsep]
  \item We propose a new framework that borrows human intuitions for designing agent capable of understanding high-level instructions, which closely integrate navigation and visual grounding in both training and inference. Specifically, the visual grounding models are pre-trained and serve as vision and language encoders for training navigation action decoder in the training phase. In inference, the action is predicted by considering logits from both the visual grounding models and the navigation decoder.
  \item We introduce two novel pre-training tasks, called Scene Grounding task and Object Grounding task, and a new Memory-augmented attentive action decoder in our framework. The pre-training tasks attempt to help the agent learn \textbf{\emph{where}} to stop and \textbf{\emph{what}} to attend to, and the action decoder effectively exploits past observations to fuse visual and textual modalities.
  \item Without bells and whistles, our method outperforms all previous methods, achieving new state-of-the-art performance on both seen and unseen environments on the REVERIE task.
\end{itemize}

\section{Related Work}
\noindent\textbf{Vision-and-Language Navigation and REVERIE.} In VLN, an agent is required to navigate to a goal location in a $3$D simulator based on fine-grained instructions. ~\cite{mattersim} proposed the Matterport3D Simulator and designed the Room-to-Room task. Then, a lot of methods have been proposed to solve this task~\cite{fried2018speaker, wang2018look, wang2019reinforced, tan2019envdrop, ke2019tactile}. On the other hand, the recently proposed REVERIE task ~\cite{qi2020reverie} is different from traditional VLN in that it requires an agent to navigate and localize target object simultaneously under the guidance of high-level instruction. The model they proposed trains the navigation model with the interactive module that works together with the object grounding model~\cite{yu2018mattnet}, in the hope that the model could learn to understand high-level instruction in a data-driven manner. However, our motivation is essentially different in that we inject commonsense knowledge prior and past memory experiences into the action policy taking into consideration the human perception in dealing with such high-level instruction navigation problems. Specifically, we introduce two pre-training tasks and a memory based action policy to make the agent become scene-intuitive. Moreover, our pre-training tasks differ from the ones proposed in ~\cite{fried2018speaker, Zhu_2020_CVPR, majumdar2020vlnbert} in that their motivation is based on the fact that the ground truth navigation path is actually hidden in the fine-grained instruction, which is not the case in high-level instruction navigation.


\noindent\textbf{Memory-based policy for navigation tasks.} Various memory models have been extensively studied for navigation agents, including unstructured memory~\cite{hochreiter1997long, piotr2017learn, Wierstra2007SolvingDM, Jaderberg2017ReinforcementLW, Mirowski2017LearningTN, Das2018EmbodiedQA}, addressable memory~\cite{Oh2016ControlOM, Parisotto2018NeuralMS}, topological memory~\cite{Savinov2018SemiparametricTM}, and metric grid-based maps~\cite{Gupta2019CognitiveMA, vln-chasing-ghosts}, etc. Unstructured memory  representations, such as LSTM memory, have been used extensively in both 2D and 3D environments. However, the issue of RNN based memory is that it does not contain context-dependent state feature storage or retrieval and does not have long time memory~\cite{vln-chasing-ghosts, zhao2020do, fang2019smt}. To address these limitations, more advaneced memory structures, such as addressable, topological, and metric based memory are proposed. In this paper, we adopt a simple adressable memory structure. The aim of using such a simple design is 1) to intentionally make it lightweight, thus reducing computational overhead, since the computational cost is important in REVERIE and our pipeline already contains heavy models; 2) to improve the performance of the overall pipeline rather than designing a more advanced memory superior to others. Besides, in VLN, the metric map memory construction requires finegrained language instruction as guidance, which is not available in our task, and building the topological memory requires pre-exploration of the environment, a technique that is certainly helpful to our agent but is beyond the discussion of this paper.

\noindent\textbf{Vision-and-Language BERT based referring expression comprehension.} Recent years have witnessed a resurgence of active research in transferrable image-text representation learning. BERT-based models ~\cite{devlin2018bert, tan2019lxmert, su2020vlbert, lu2020multitask, chen2020uniter, lu2019vilbert} have achieved superior performance over multiple vision-and-language tasks by transferring the pre-trained model on large aligned image-text pairs to other downstream tasks. In BERT-based VLN, the most related agents to ours are ~\cite{hao2020prevalent} and ~\cite{majumdar2020vlnbert}. ~\cite{hao2020prevalent} treats VLN as a vision-and-language alignment task and utilizes a pre-trained vision-and-language BERT model to predict action sequence while ~\cite{majumdar2020vlnbert} formulates VLN as an instruction and path alignment task and adopts a pre-trained vision-and-language BERT model to find the best candidate path that matches to the instruction given. However, our work differs from others in that we propose a generalized pipeline that mimics human intuitions to solve the high-level instruction navigation task where vision-and-language BERT model is a building block which can be customized to other vision-language alignment block. Experimental results show that the main performance gain comes from our proposed pipeline.

\begin{figure*}[ht]
  \centering
  \includegraphics[width=\linewidth]{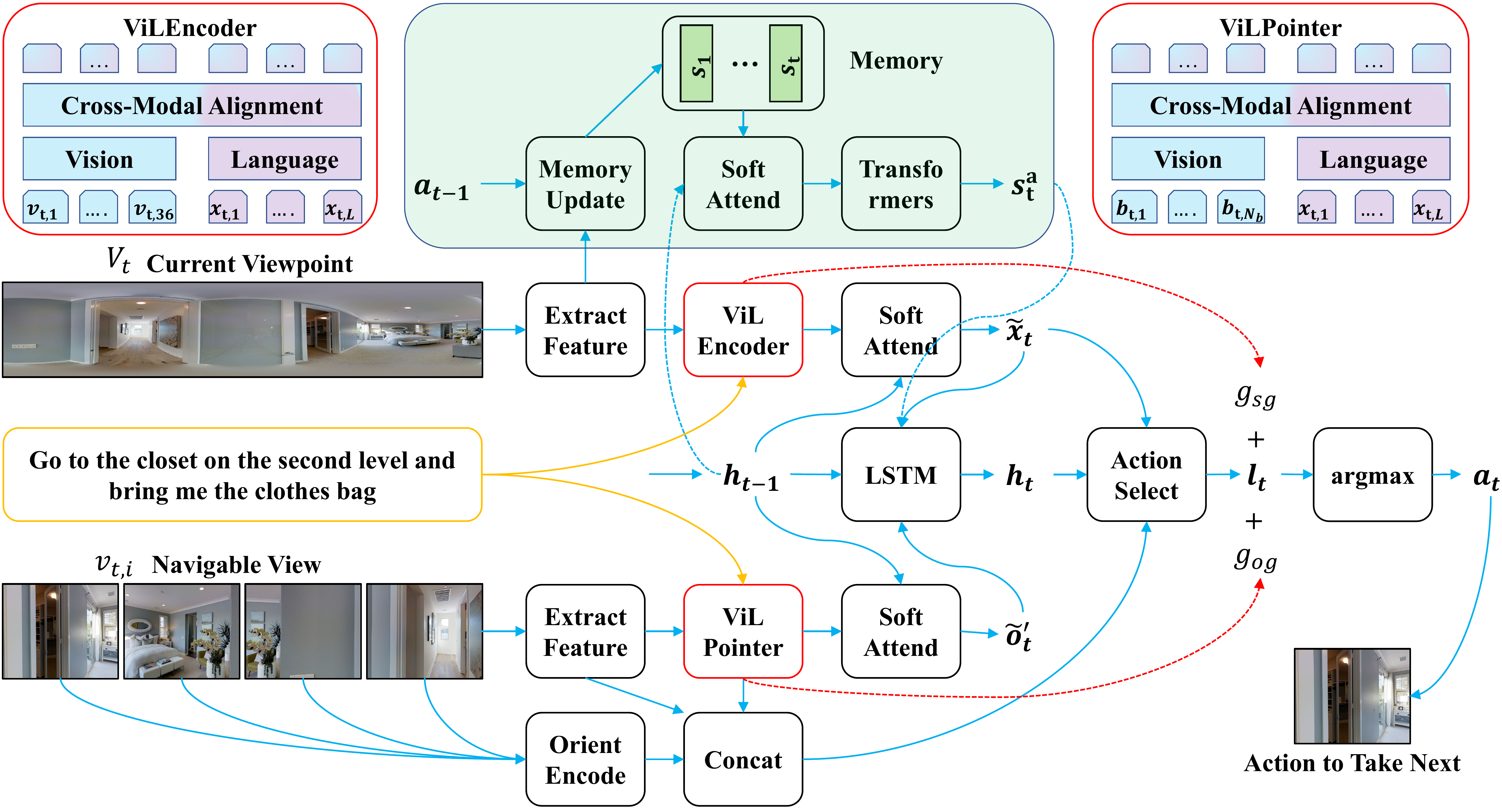}
  \caption{The overall pipeline of our method. The green part of the figure denotes the memory module where current viewpoint feature $\boldsymbol{V}_{t}$ and previous action feature $\boldsymbol{a}_{t-1}$ are embedded and stored in the Memory. $Transformer$ blocks are used to generate $\boldsymbol{s}_{t}^{a}$. The red rectangles represent two pre-trained models, namely Scene Grounding model and Object Grounding model. $ViLEncoder$ consists of $ViLBERT$ and $BiLSTM$ and $ViLPointer$ is $ViLBERT$ trained on viewpoint-based object grounding task. At each time step $t$, the agent perceives the instruction with viewpoint features and object features simultaneously. Action prediction is made by the Action Select part where an attentive structure is applied. The final action is generated by considering scene grounding score $g_{sg}$, object grounding score $g_{og}$ and action logit $\boldsymbol{l}_{t}$. The dashed dot lines are used only for illustration purposes.}
  \label{Fig:pipeline}
\end{figure*}

\section{Method}
\noindent In the REVERIE task, an agent placed at a starting location navigates to the target location to localize an object specified by a high-level instruction. To carry out this difficult task, we propose a novel pipeline that contains a scene grounding model, an object grounding model, and a memory-based action decoder. We make two claims of our design choice: first, to better grasp the semantics of high-level instructions, we choose ViLBERT model as our basic building block to serve as vision-and-language encoder; second, since scene grounding task and object grounding task are two essentially different tasks, we do not share the basic building blocks for these two tasks. In general, we decompose our method into two stages, as shown in Fig.~\ref{Fig:pipeline}, namely the pre-training stage and the action decoding stage. In the following sections, we first introduce the pre-training tasks; then we illustrate the memory-based attentive action decoder and finally, the loss function used to train the agent.

\subsection{ViLBERT introduction}
\begin{figure}[ht]
  \centering
  \includegraphics[width=\linewidth]{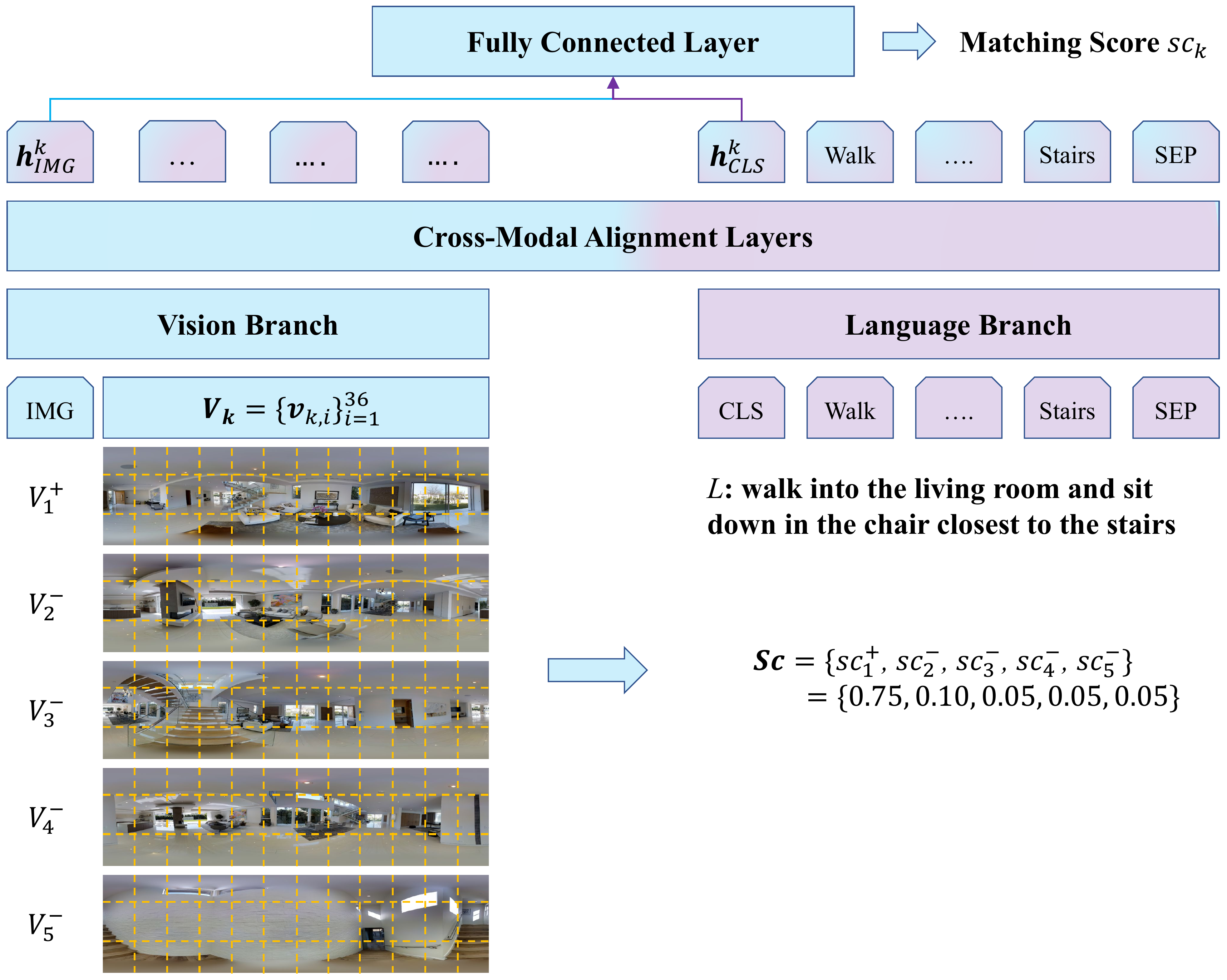}
  \caption{The pipeline of the Scene Grounding Task. We formulate this task as a $5$-way multiple choice problem. Each $(L, ResNet(V_{k}))$ pair is sent to the $ViLBERT$ model separately to generate alignment score $sc_{k}$. The panorama viewpoint image here denotes the discretized $36$ view images in a viewpoint. We mark the beginning of the image sequence with a special token $IMG$ and the language with $CLS$.}
  \label{Fig:SceneGroundingTask}
\end{figure}

\noindent In this section, we briefly introduce the input and output arguments of a ViLBERT model~\cite{lu2019vilbert} as shown in Fig.~\ref{Fig:SceneGroundingTask}. A ViLBERT model is a BERT-based model that consists of two input streams, vision encoding stream and language encoding stream, followed by a cross-modal alignment Transformer block. The inputs to ViLBERT model are sequence of words and visual features respectively and the outputs are corresponding encoded word sequence features as well as visual sequence features. We use ViLBERT as our base model (basic building block) for the Scene Grounding task and the Object Grounding task. In Scene Grounding task, a panorama viewpoint image is discretized into $36$ view images and the inputs are sequence of words in the instruction and $36$ mean-pooled features extracted from $36$ view images by a ResNet-$152$ CNN pre-trained on ImageNet~\cite{krizhevsky2012imagenet}. In Object Grounding task, the inputs are sequence of words in the instruction and all annotated bounding boxes features extracted by Mask R-CNN ~\cite{he2017mask} in a target viewpoint.

\subsection{Overview of the proposed method}
\noindent \textbf{Settings.} To formalize the task, we denote a given high-level instruction as $L = \left \{ l_{k}\right \}_{k=1}^{N_{l}}$  where $N_{l}$ is the number of words in the instruction $L$ and a set of viewpoints as $\nu = \left \{ V_{k}\right \}_{k=1}^{N_{v}}$ where $N_{v}$ is the number of viewpoints in the environment. At each time step $t$, the agent observes a panoramic view $V_{t}$, a few navigable views $O_{t}$ and a set of annotated bounding boxes $B_{t}$. The panoramic view is discretized into $36$ single views by perspective projections, each of which is a $640\times480$ size image with field of view set to $60$ degrees, and is denoted by $V_{t}=\left \{v_{t,i} \right \}_{i=1}^{36}$. $O_{t} = \left \{v_{t,i} \right \}_{i=1}^{N_{o}} \subseteq V_{t}$ where $N_{o}$ is the maximum navigable directions at a viewpoint $V_{t}$. Each $v_{t, i}$ is represented as $\boldsymbol{v}_{t, i} = ResNet(v_{t,i})$. Thus, $\boldsymbol{V}_{t} = \left \{\boldsymbol{v}_{t,i} \right \}_{i=1}^{36}$. Besides, the set of annotated bounding boxes at viewpoint $V_{t}$ is denoted by $B_{t} = \left \{b_{t, i} \right \}_{i=1}^{N_{b}}$ where $N_{b}$ is the number of bounding boxes. Mask R-CNN~\cite{he2017mask} is used to extract bounding boxes features $\boldsymbol{B}_{t} = \left \{\boldsymbol{b}_{t, i} \right \}_{i=1}^{N_{b}}$, where $\boldsymbol{b}_{t,i} = MRCNN(b_{t,i})$.

\textbf{Stage 1(a): Scene Grounding Task.} We formulate the task as finding a viewpoint that best matches to a high-level instruction $L$ in a set of candidate viewpoints $\nu_{s}$. $\nu_{s} = \left \{ V_{k} | V_{k} \in \nu \right \} \subseteq \nu$. Concretely, we define a mapping function $g_{sg}(,)$ that maps $(L, V_{k})$ to a matching score. The formula is defined as follows,
    \begin{equation}
    \begin{aligned}
    	V_{k}^{\star}&=\mathop{\arg\max}_{V_{k} \in \nu_{s}}g_{sg}(L, ResNet(V_{k}))
    \end{aligned}
    \end{equation}

\textbf{Stage 1(b): Object Grounding Task.} The goal of this task is to identify the best matching object among a set of candidate objects located at a target viewpoint. We denote $V_{T}$ as a target viewpoint and its corresponding annotated bounding boxes set is $B_{T}$. We define another compatibility matching function $g_{og}(,)$ that produce matching scores for all objects with a high-level instruction $L$. Thus, the problem is defined as follows,
    \begin{equation}
    \begin{aligned}
    	b_{T,i}^{\star}&=\mathop{\arg\max}_{b_{T, i} \in B_{T}}g_{og}(L, MRCNN(b_{T, i}))
    \end{aligned}
    \end{equation}

\textbf{Stage 2: Memory-augmented action decoder.} To mitigate the memory problem presented in previous section, a scene memory structure $\boldsymbol{M}_{t}$ is implemented to store the embedded observation and previous action at each time step $t$. The memory is updated by,
    \begin{equation}
    \begin{aligned}
        \tilde{\boldsymbol{v}}_{t} &= softmax(\boldsymbol{V}_{t}(\boldsymbol{W}_{1}\boldsymbol{h}_{t-1}))^{T}\boldsymbol{V}_{t} \\
        \boldsymbol{s}_{t} &= FC([\boldsymbol{a}_{t-1}, \tilde{\boldsymbol{v}}_{t}]), \\
    	\boldsymbol{M}_{t}&=Update(\boldsymbol{M}_{t-1}, \boldsymbol{s}_{t})
    \end{aligned}
    \end{equation}
where $\boldsymbol{s}_{t}$ is current state representation; $\tilde{\boldsymbol{v}}_{t}$ is attentive visual feature;$\boldsymbol{h}_{t-1}$ and $\boldsymbol{a}_{t-1}$ are last time step hidden state and action embedding respectively;$\boldsymbol{W}_{1}\in\mathbb{R}^{2048 \times D_{h}}$ is a trainable parameter. $FC$ stands for fully connected layer. The $Update$ operation appends $\boldsymbol{s}_{t}$ to $\boldsymbol{M}_{t}$. $\boldsymbol{V}_{t}\in\mathbb{R}^{36\times2048}, \tilde{\boldsymbol{v}}_{t}\in\mathbb{R}^{1\times2048}, \boldsymbol{a}_{t-1}\in\mathbb{R}^{1\times3200}, \boldsymbol{s}_{t}\in\mathbb{R}^{1 \times D_{h}}, \boldsymbol{h}_{t}\in\mathbb{R}^{D_{h}\times1}, \boldsymbol{M}_{t}\in\mathbb{R}^{t \times D_{h}}$.

\subsection{Scene Grounding Task}
\noindent The goal of this task it to help the agent infer where the target location is. Given a high-level instruction, \emph{``Bring me the \textbf{\textit{jeans}} that are hanging up in the \textbf{\textit{closet}} to the right''}, humans first locate the \textbf{\textit{where}} information, the key word \textbf{\textit{closet}}, by capturing the semantics of the instruction according to the language context and commonsense knowledge and then form an overview of the appearance of the \textbf{\textit{closet}} in mind; then, humans navigate to the target location by consistently matching the \textbf{\textit{closet}} appearance in mind with current scene. In fact, humans have gradually formed intuitions towards the understanding of scenes, instructions and tasks in life. For language instructions in relatively simple life scenes that do not involve complex reasoning, they usually directly merge the above two processes for direct perception and understanding. We call this process as context-driven scene perception. In this section, we propose Scene Grounding task to imitate such human behavior.

Based on the observation, we believe that a model that could evaluate the alignment between an instruction and a viewpoint is able to localize the target viewpoint. Therefore, to implement this idea, we create a dataset from the REVERIE training set and fine-tune a ViLBERT model on the dataset. Specifically, we adopt a $5$-way multiple choice setting. We eliminate subscript for simplicity concern. Given an instruction $L$, we sample $5$ viewpoints $\left \{ V_{1}^{+}, V_{2}^{-}, V_{3}^{-}, V_{4}^{-}, V_{5}^{-} \right \}$, out of which only one is aligned to the instruction (or in other words, positive). In detail, we choose the ending viewpoint in the ground-truth training path as $V_{1}^{+}$, the second last viewpoint along the ground-truth path as $V_{2}^{-}$ which is a hard negative sample and random sample $V_{3}^{-}, V_{4}^{-}$ from the rest of the viewpoints along the path, and $V_{5}^{-}$ from other path . Then, we run the ViLBERT model on each of the $(L, V_{k})$ pair. As is shown in Fig.~\ref{Fig:SceneGroundingTask}, the output tokens $CLS$ and $IMG$ encode instruction representation $\boldsymbol{h}_{CLS}$ as well as viewpoint representation $\boldsymbol{h}_{IMG}$ respectively. We define the matching scores as $\boldsymbol{Sc}$ and train the model with cross entropy loss $\mathcal{L}_{sr}$.
    \begin{equation}
    \begin{aligned}
    	\boldsymbol{Sc}&=\left \{sc_{1}, sc_{2}, sc_{3}, sc_{4}, sc_{5} \right \} \\
        sc_{k}&=g_{sg}(L, ResNet(V_{k}))=\boldsymbol{W}_{2}(\boldsymbol{h}_{CLS}^{k} \odot \boldsymbol{h}_{IMG}^{k})\\
        \mathcal{L}_{sr}&=CELoss(softmax(\boldsymbol{Sc}), \mathbb{I}(V_{1}^{+}))
    \end{aligned}
    \end{equation}
where $\boldsymbol{W}_{2}\in\mathbb{R}^{1\times1024}$ is a trainable parameter and $\mathbb{I}(.)$ is indicator function. $\boldsymbol{h}_{CLS}^{k}\in\mathbb{R}^{1024\times1}, \boldsymbol{h}_{IMG}^{k}\in\mathbb{R}^{1024\times1}$ are the encoded language and visual representations of the language and vision encoding streams from our pre-trained ViLBERT model for $k$th $(L, V_{k})$ pair respectively.

\subsection{Object Grounding Task}
\noindent The aim of this task is to help the agent learn what to attend to. For each ground-truth target viewpoint $V_{T}$, we formulate this task as finding the best bounding box $b_{T, i}^{\star}$ in bounding boxes set $B_{T}$ given $(L, B_{T})$ pair. A straightforward method to implement this idea is to construct a single image based grounding task, where each training sample consists of instruction $L$ and a subset of bounding boxes in $B_{T}$ that belong to view $v_{T, i}$. However, according to our experiment, this strategy produces moderate performance since objects in $3$D space could span multiple views in corresponding projected $2$D image space. The cross-image objects relationships in each viewpoint are not well captured by the model. Therefore, we propose a two-stage training strategy, namely a single image based grounding and a viewpoint based object grounding. In single image grounding, we fine-tune the ViLBERT model from ~\cite{lu2020multitask, lu2019vilbert} on the aforementioned single image grounding dataset where each training sample is $(L, B_{v_{T, i}})$ (all annotated bounding boxes in $v_{T, i}$ are collected) and $B_{v_{T, i}}\subset B_{T}$; then, we further fine-tune trained model on a new viewpoint based object grounding dataset. Concretely, each training sample in the viewpoint based dataset is a $(L, B_{T})$ pair (all annotated bounding boxes in $v_{T}$ are collected) and the corresponding label is a vector containing $0$s and $1$s where $1$ indicates the IoU of a bounding box with the target bounding box is higher than $0.5$. In inference, we represent an object score as the averaged scores from all bounding boxes that share the same object id at a viewpoint that the agent stops.

\subsection{Action Decoder}
\noindent With the pre-trained grounding models, the action decoder generally adopts Encoder-Decoder structure to produce action prediction. Specifically, the Scene Grounding model is accompanied by a $BiLSTM$ network to construct a vision and language grounding encoder $ViLEncoder$ and the Object Grounding model is formulated as an object level grounding encoder $ViLPointer$. The inputs to action decoder are $L$, $\boldsymbol{B}_{t}$ and $\boldsymbol{V}_{t}$ and it outputs predicted action distribution $\boldsymbol{l}_{t}$.

\textbf{First}. At each time step $t$, to perceive current scene and instruction, we obtain $\tilde{\boldsymbol{x}}_{t}$ by grounding $L$ with $\boldsymbol{V}_{t}$ through $ViLEncoder$ and then selecting the fused language sequence as output. The formula is defined as follows,
    \begin{equation}
    \begin{aligned}
        \boldsymbol{X}_{t}&=ViLEncoder(L, \boldsymbol{V}_{t}) \\
        &=BiLSTM(ViLBERT(L, \boldsymbol{V}_{t})) \\
        \tilde{\boldsymbol{x}}_{t}&=softmax(\boldsymbol{X}_{t}(\boldsymbol{W}_{3}\boldsymbol{h}_{t-1}))^{T}\boldsymbol{X}_{t}
    \end{aligned}
    \end{equation}
where $\boldsymbol{W}_{3}\in\mathbb{R}^{1024 \times D_{h}}$ is a trainable parameter and $\boldsymbol{X}_{t}$ is encoded language feature taking current scene $\boldsymbol{V}_{t}$ into consideration. $\boldsymbol{X}_{t}\in\mathbb{R}^{N_{l}\times1024}, \tilde{\boldsymbol{x}}_{t}\in\mathbb{R}^{1\times1024}, \boldsymbol{h}_{t-1}\in\mathbb{R}^{D_{h}\times1}$.

\textbf{Second}. To decide which navigable direction to go next, we perform object level referring expression comprehension. The object level referring comprehension helps the agent infer whether a navigable view $v_{t, i}$ contains possible target object. In particular, the set of bounding boxes in view $v_{t, i}$ is denoted by $\hat{B}_{t, i}=\left \{ b_{t, k} | b_{t, k} \in B_{t}, Inside(b_{t, k}, v_{t, i})=1\right \}$ where $Inside(,)$ function decides whether $b_{t, k}$ is inside view $v_{t, i}$. $ViLPointer$ is $ViLBERT$ pre-trained on the Object Grounding task and we select the fused bounding boxes features as the output. Then,
    \begin{equation}
    \begin{aligned}
        \boldsymbol{F}_{t, i}&=ViLPointer(L, MRCNN(\hat{B}_{t, i})) \\
        \tilde{\boldsymbol{v}}_{t, i}&=g_{top-k}(\boldsymbol{F}_{t, i})
    \end{aligned}
    \end{equation}
where $\boldsymbol{F}_{t, i}$ is the set of aligned bounding boxes features at view $v_{t, i}$ and $g_{top-k}(,)$ selects top-$k$ aligned bounding boxes and averages the corresponding aligned bounding boxes features from $\boldsymbol{F}_{t, i}$ to produce view comprehension $\tilde{\boldsymbol{v}}_{t, i}\in\mathbb{R}^{1\times1024}$.

\textbf{Third}. We define the representation of each navigable view as $\boldsymbol{v}_{t, i}'$:
    \begin{equation}
    \begin{aligned}
        \boldsymbol{v}_{t, i}'&=[\boldsymbol{v}_{t, i}, (\cos\theta_{t, i}, \sin\theta_{t, i}, \cos\phi_{t, i}, \sin\phi_{t, i}), \tilde{\boldsymbol{v}}_{t, i}]
    \end{aligned}
    \end{equation}
where the agent's current orientation $(\theta_{t, i}, \phi_{t, i})$ represents the angles of heading and elevation and is tiled $32$ times according to ~\cite{fried2018speaker}. $(\cos\theta_{t, i}, \sin\theta_{t, i}, \cos\phi_{t, i}, \sin\phi_{t, i})\in\mathbb{R}^{1\times128}$ and $\boldsymbol{v}'_{t, i}\in\mathbb{R}^{1\times3200}$. The set of navigable view representation is denoted as $\boldsymbol{O}'_{t} = \left \{ \boldsymbol{v}'_{t, i}\right \}_{i=1}^{N_{o}}$. The grounded navigable visual representation $\tilde{\boldsymbol{o}}'_{t}$ is represented as follows:
    \begin{equation}
    \begin{aligned}
        \tilde{\boldsymbol{o}}'_{t} = softmax(g(\boldsymbol{O}'_{t})(\boldsymbol{W}_{4}\boldsymbol{h}_{t-1}))^{T}g(\boldsymbol{O}'_{t})
    \end{aligned}
    \end{equation}
where $\boldsymbol{W}_{4}\in\mathbb{R}^{1024 \times D_{h}}$ is a trainable parameter and $g(,)$ is a number of Fully Connected layers accompanied by ReLU nonlinearities. $\tilde{\boldsymbol{o}}'_{t}\in\mathbb{R}^{1\times1024}, \boldsymbol{O}'_{t}\in\mathbb{R}^{N_{o}\times3200}$.

\textbf{Fourth}. The new context hidden state $\boldsymbol{h}_{t}$ is updated by a LSTM layer taking as input the grounded text $\tilde{\boldsymbol{x}}_{t}$ and navigable view features $\tilde{\boldsymbol{o}}'_{t}$ as well as the current state representation feature $\boldsymbol{s}_{t}^{a}$.
    \begin{equation}
    \begin{aligned}
        (\boldsymbol{h}_{t}, \boldsymbol{c}_{t}) = LSTM([\tilde{\boldsymbol{x}}_{t}, \tilde{\boldsymbol{o}}'_{t}, \boldsymbol{s}_{t}^{a}], (\boldsymbol{h}_{t-1}, \boldsymbol{c}_{t-1}))
    \end{aligned}
    \end{equation}
where $\boldsymbol{s}_{t}^{a}$ is memory augmented current state representation and is defined as,
    \begin{equation}
    \begin{aligned}
        \boldsymbol{M}_{t}^{a}&=[Transformer(\boldsymbol{M}_{t}, \boldsymbol{M}_{t})]_{\times N_{mem}} \\
        \boldsymbol{s}_{t}^{a}&=[Transformer(\boldsymbol{s}_{t}, \boldsymbol{M}_{t}^{a})]_{\times N_{state}} \\
    \end{aligned}
    \end{equation}
where $N_{mem}$ and $N_{state}$ are number of memory transformer blocks used and number of state transformer blocks used respectively. $\boldsymbol{s}_{t}^{a}\in\mathbb{R}^{1 \times D_{h}}, \boldsymbol{M}_{t}^{a}\in\mathbb{R}^{t \times D_{h}}$. $Transformer$ is the standard version Transformer block from ~\cite{google2017attention}.

\textbf{Finally}. The action logit $\boldsymbol{l}_{t}$ is computed in an attentive manner.
    \begin{equation}
    \begin{aligned}
        l_{t, i} = g(\boldsymbol{O}'_{t, i})(\boldsymbol{W}_{5}[\boldsymbol{h}_{t}, \tilde{\boldsymbol{x}}_{t}])
    \end{aligned}
    \end{equation}
where $\boldsymbol{W}_{5}\in\mathbb{R}^{1024\times(1024+D_{h})}$ is a trainable parameter and $\boldsymbol{l}_{t}\in\mathbb{R}^{N_{o}\times1}$. In training stage, $a_{t} = Categorical(\boldsymbol{l}_{t})$ is selected based on categorical policy and in inference stage, it is selected by $a_{t} = \arg max(\boldsymbol{l}_{t})$. Action embedding is selected based on $\boldsymbol{a}_{t} = \boldsymbol{O}'_{t}[a_{t}]$.

\subsection{Inference}
\noindent We propose to use a combined logit $\sum_{\tau=0}^{t}\boldsymbol{l}_{\tau} + g_{og}^{\tau} + g_{sg}^{\tau}$ that sums action logits, object grounding logits and scene grounding logits to perform navigation, where $g_{og}^{\tau}$ and $g_{sg}^{\tau}$ denote object grounding score and scene grounding score at time step $\tau$ respectively.
Experimental results indicate that our strategy shortens the search trajectories while maintaining a good success rate. The final output bounding box is obtained by running $ViLPointer$ at the stop viewpoint that the agent predicts.

\subsection{Loss Functions}
\noindent To train the agent, we use a mixture of Imitation Learning (IL) and Reinforcement Learning (RL) to supervise the training. Specifically, In IL, at each time step, we allow the agent to learn to imitate the teacher action by using a cross entropy loss $\mathcal{L}_{ce}$ and a mean squared error loss $\mathcal{L}_{pm}$ for progress monitor ~\cite{ma2019selfmonitoring}. In RL, we follow the idea of ~\cite{tan2019envdrop} and allow the agent to learn from rewards. If the agent stops within $3$ meters near the target viewpoint, a positive reward $+3$ is assigned at the final step; otherwise a negative reward $-3$ is given.
    \begin{equation}
    \begin{aligned}
        \mathcal{L}_{final}&=\alpha\mathcal{L}_{ce}+\beta\mathcal{L}_{pm}+\gamma\mathcal{L}_{RL} \\
        \mathcal{L}_{ce}&=-\sum_{t=1}^{T}y_{t}^{\star}\log(l_{t, \star}) \\
        \mathcal{L}_{pm}&=-\sum_{t=1}^{T}(y_{t}^{pm} - p_{t}^{pm})^{2}
    \end{aligned}
    \end{equation}
where $y_{t}^{\star}$ is the teacher action at step $t$; $y_{t}^{pm}\in[0, 1]$ is the shortest normalized distance from current viewpoint to the target viewpoint; $p_{t}^{pm}$ is the predicted progress; $\alpha$, $\beta$ and $\gamma$ are all set to $1$.

\section{Experiments}
\noindent In the REVERIE dataset, the training set contains $59$ scenes and $10466$ instructions over $2353$ objects; the val seen split consists of $53$ scenes and $1371$ instructions over $428$ objects and the val unseen split include $10$ scenes and $3573$ instructions over $525$ objects. The test set contains $16$ scenes and $6292$ instructions over $834$ objects. In this section, we conduct extensive evaluation and analysis of the effectiveness of our proposed components.

\begin{table*}[t]
\centering
\caption{Ablation Study experiments performed to verify the effectiveness of the proposed method. In different ablation study block, the best performing result is marked in \underline{\textbf{bold}}.}
\resizebox{\linewidth}{!}{
\begin{tabular}{c|c|cccccccc|cccc|cc|cccc|cc}
\hline
\multirow{3}{*}{Experiments} & \multirow{3}{*}{ID} & \multicolumn{8}{c|}{Methods} & \multicolumn{6}{c|}{Val Seen} & \multicolumn{6}{c}{Val Unseen} \\
\cline{3-22}
 &  & \multicolumn{4}{c|}{Encoder} & \multicolumn{2}{c|}{Pointer} & \multicolumn{2}{c|}{Policy} & \multicolumn{4}{c|}{Nav. Acc.} & \multicolumn{1}{c|}{\multirow{2}{*}{\begin{tabular}[c]{@{}c@{}}RGS$\uparrow$\\ \end{tabular}}} & \multirow{2}{*}{\begin{tabular}[c]{@{}c@{}}RG\\SPL$\uparrow$ \end{tabular}} & \multicolumn{4}{c|}{Nav. Acc.} & \multicolumn{1}{c|}{\multirow{2}{*}{RGS$\uparrow$ }} & \multirow{2}{*}{\begin{tabular}[c]{@{}c@{}}RG\\SPL$\uparrow$ \end{tabular}} \\
\cline{3-14}\cline{17-20}
 &  & \multicolumn{1}{c|}{$L_{enc}$} & \multicolumn{1}{c|}{$Bert_{enc}$} & \multicolumn{1}{c|}{$ViLRaw_{enc}$} & \multicolumn{1}{c|}{$ViL_{enc}$} & \multicolumn{1}{c|}{$MN_{ptr}$ } & \multicolumn{1}{c|}{$ViL_{ptr}$} & \multicolumn{1}{c|}{$C_{pol}$} & $MA_{pol}$  & Succ.$\uparrow$  & OSucc.$\uparrow$  & SPL$\uparrow$  & Length$\downarrow$  & \multicolumn{1}{c|}{} &  & Succ.$\uparrow$  & OSucc.$\uparrow$  & SPL$\uparrow$  & Length$\downarrow$  & \multicolumn{1}{c|}{} &  \\
\hline
\multirow{8}{*}{\begin{tabular}[c]{@{}c@{}}Component\\Effectiveness \end{tabular}} & 1 & $\surd$ &  &  &  & $\surd$ &  & $\surd$ &  & 50.53 & 55.17 & 45.50 & 16.35 & 31.97 & 29.66 & 14.40 & 28.20 & 7.19 & 45.28 & 7.84 & 4.67 \\
 & 2 &  & $\surd$ &  &  & $\surd$ &  & $\surd$ &  & 54.18 & 58.68 & 48.99 & \uline{\textbf{12.46}} & 33.87 & 21.23 & 18.66 & 29.51 & 10.44 & \uline{\textbf{32.95}} & 11.13 & 6.32 \\
 & 3 &  &  & $\surd$ &  & $\surd$ &  & $\surd$ &  & 33.73 & 39.14 & 30.72 & 14.56 & 23.82 & 21.94 & 15.22 & 31.64 & 8.44 & 42.62 & 8.89 & 4.84 \\
 & 4 &  &  & $\surd$ &  &  & $\surd$ & $\surd$ &  & 39.00 & 43.85 & 35.00 & 13.71 & 28.95 & 25.98 & 13.80 & 31.33 & 8.21 & 37.31 & 9.17 & 5.54 \\
 & 5 &  &  & $\surd$ &  & $\surd$ &  &  & $\surd$ & 37.32 & 43.08 & 31.71 & 18.29 & 24.88 & 21.70 & 19.06 & 44.39 & 7.10 & 79.88 & 11.08 & 4.17 \\
 & 6 &  &  &  & $\surd$ & $\surd$ &  & $\surd$ &  & 56.36 & 60.93 & 52.24 & 13.21 & 36.33 & 33.92 & 21.61 & 31.98 & 12.21 & 36.05 & 13.21 & 7.31 \\
 & 7 &  &  &  & $\surd$ &  & $\surd$ & $\surd$ &  & 54.25 & 56.08 & 50.49 & 13.56 & 39.56 & 37.16 & 26.98 & 37.86 & 13.70 & 42.50 & 17.32 & 8.71 \\
 & 8 &  &  &  & $\surd$ &  & $\surd$ &  & $\surd$ & \uline{\textbf{59.52}}  & \uline{\textbf{64.23}}  & \uline{\textbf{55.30}}  & 14.00 & \uline{\textbf{43.57}}  & \uline{\textbf{40.42}}  & \uline{\textbf{28.17}}  & \uline{\textbf{40.41}}  & \uline{\textbf{14.77}}  & 43.12 & \uline{\textbf{19.60}}  & \uline{\textbf{10.27}}  \\
\hline
\multirow{5}{*}{\begin{tabular}[c]{@{}c@{}}Memory\\Blocks\\($N_{mem}$, $N_{state}$) \end{tabular}} & 9 & \multicolumn{8}{c|}{(1, 1)} & 55.24 & 58.61 & 52.29 & \uline{\textbf{12.42}}  & 40.90 & 38.76 & 28.97 & 39.56 & 13.28 & 44.10 & 20.51 & 9.19 \\
 & 10 & \multicolumn{8}{c|}{(3, 3)} & \uline{\textbf{61.91}}  & \uline{\textbf{65.85}}  & \uline{\textbf{57.08}}  & 13.61 & \uline{\textbf{45.96}}  & \uline{\textbf{42.65}}  & \uline{\textbf{31.53}}  & \uline{\textbf{44.67}}  & \uline{\textbf{16.28}}  & \uline{\textbf{41.53}}  & \uline{\textbf{22.41}}  & \uline{\textbf{11.56}}  \\
 & 11 & \multicolumn{8}{c|}{(5, 5)} & 60.01 & 63.38 & 54.99 & 17.44 & 44.69 & 41.10 & 25.84 & 38.20 & 13.09 & 44.00 & 18.23 & 9.19 \\
 & 12 & \multicolumn{8}{c|}{(7, 7)} & 57.27 & 62.26 & 52.78 & 13.96 & 42.66 & 39.38 & 23.66 & 35.61 & 11.67 & 45.73 & 16.79 & 8.43 \\
 & 13 & \multicolumn{8}{c|}{(9, 9)} & 57.06 & 60.15 & 53.35 & 14.16 & 42.38 & 39.67 & 28.15 & 39.45 & 14.92 & 41.53 & 19.54 & 10.13 \\
\hline
\multirow{4}{*}{\begin{tabular}[c]{@{}c@{}}Logit\\Fusion \end{tabular}} & 14 & \multicolumn{8}{c|}{$\boldsymbol{l}_{t}$ } & 60.92 & 65.78 & 56.14 & 15.28 & 45.61 & 42.19 & \uline{\textbf{32.35}}  & \uline{\textbf{49.08}}  & 14.74 & 60.89 & 22.35 & 10.54 \\
 & 15 & \multicolumn{8}{c|}{$\boldsymbol{l}_{t}$+$g_{sg}$ } & 61.49 & 65.78 & 56.72 & 13.67 & 45.47 & 42.31 & 31.20 & 47.80 & 15.90 & 45.82 & 21.68 & 11.08 \\
 & 16 & \multicolumn{8}{c|}{$\boldsymbol{l}_{t}$+$g_{og}$ } & 61.14 & 65.77 & 55.21 & 16.82 & 44.48 & 40.04 & 32.12 & 46.54 & 15.73 & 52.14 & 21.98 & 11.02 \\
 & 17 & \multicolumn{8}{c|}{$\boldsymbol{l}_{t}$+$g_{sg}$+$g_{og}$ } & \uline{\textbf{61.91}}  & \uline{\textbf{65.85}}  & \uline{\textbf{57.08}}  & \uline{\textbf{13.61}}  & \uline{\textbf{45.96}}  & \uline{\textbf{42.65}}  & 31.53 & 44.67 & \uline{\textbf{16.28}}  & \uline{\textbf{41.53}}  & \uline{\textbf{22.41}}  & \uline{\textbf{11.56}}  \\
\hline
\end{tabular}}
\label{tab:ablation}
\end{table*}

\begin{table*}[t]
\centering
\caption{Comparison with state-of-the-art methods on the REVERIE task. The best performing result is marked in \underline{\textbf{bold}}.}
\resizebox{\linewidth}{!}{
\begin{tabular}{c||cccc|cc||cccc|cc||cccc|cc}
\hline
\multirow{3}{*}{Methods} & \multicolumn{6}{c||}{Val Seen} & \multicolumn{6}{c||}{Val Unseen} & \multicolumn{6}{c}{Test (Unseen)} \\
\cline{2-19}
 & \multicolumn{4}{c|}{Nav. Succ.} & \multicolumn{1}{c|}{\multirow{2}{*}{RGS$\uparrow$}} & \multirow{2}{*}{\begin{tabular}[c]{@{}c@{}}RG\\SPL\end{tabular}$\uparrow$} & \multicolumn{4}{c|}{Nav. Succ.} & \multicolumn{1}{c|}{\multirow{2}{*}{RGS$\uparrow$}} & \multirow{2}{*}{\begin{tabular}[c]{@{}c@{}}RG\\SPL\end{tabular}$\uparrow$} & \multicolumn{4}{c|}{Nav. Succ.} & \multicolumn{1}{c|}{\multirow{2}{*}{RGS$\uparrow$}} & \multirow{2}{*}{\begin{tabular}[c]{@{}c@{}}RG\\SPL\end{tabular}$\uparrow$} \\
\cline{2-5}\cline{8-11}\cline{14-17}
 & Succ.$\uparrow$ & OSucc.$\uparrow$ & SPL$\uparrow$ & Length$\downarrow$ & \multicolumn{1}{c|}{} &  & Succ.$\uparrow$ & OSucc.$\uparrow$ & SPL$\uparrow$ & Length$\downarrow$ & \multicolumn{1}{c|}{} &  & Succ.$\uparrow$ & OSucc.$\uparrow$ & SPL$\uparrow$ & Length$\downarrow$ & \multicolumn{1}{c|}{} &  \\
\hline
RCM~\cite{wang2019reinforced} + MattNet& 23.33 & 29.44 & 21.82 & 10.70 & 16.23 & 15.36 & 9.29 & 14.23 & 6.97 & 11.98 & 4.89 & 3.89 & 7.84 & 11.68 & 6.67 & 10.60 & 3.67 & 3.14 \\
SelfMonitor~\cite{ma2019selfmonitoring} + MattNet & 41.25 & 43.29 & 39.61 & \underline{\textbf{7.54}} & 30.07 & 28.98 & 8.15 & 11.28 & 6.44 & \underline{\textbf{9.07}} & 4.54 & 3.61 & 5.80 & 8.39 & 4.53 & \underline{\textbf{9.23}} & 3.10 & 2.39 \\
FAST-short~\cite{ke2019tactile} + MattNet & 45.12 & 49.68 & 40.18 & 13.22 & 31.41 & 28.11 & 10.08 & 20.48 & 6.17 & 29.70 & 6.24 & 3.97 & 14.18 & 23.36 & 8.74 & 30.69 & 7.07 & 4.52 \\
REVERIE~\cite{qi2020reverie} & 50.53 & 55.17 & 45.50 & 16.35 & 31.97 & 29.66 & 14.40 & 28.20 & 7.19 & 45.28 & 7.84 & 4.67 & 19.88 & 30.63 & 11.61 & 39.05 & 11.28 & 6.08 \\
Human & - & - & - & - & - & - & - & - & - & - & - & - & 81.51 & 86.83 & 53.66 & 21.18 & 77.84 & 51.44 \\
\hline
Ours & \underline{\textbf{61.91}} & \underline{\textbf{65.85}} & \underline{\textbf{57.08}} & 13.61 & \underline{\textbf{45.96}} & \underline{\textbf{42.65}} & \underline{\textbf{31.53}} & \underline{\textbf{44.67}} & \underline{\textbf{16.28}} & 41.53 & \underline{\textbf{22.41}} & \underline{\textbf{11.56}} & \underline{\textbf{30.8}} & \underline{\textbf{44.56}} & \underline{\textbf{14.85}} & 48.61 & \underline{\textbf{19.02}} & \underline{\textbf{9.20}} \\
\hline
\end{tabular}}
\label{tab:sota}
\end{table*}

\subsection{Evaluation Metrics}
\noindent Following ~\cite{qi2020reverie}, we evaluate the performance of the model based on REVERIE Success Rate (RGS) and REVERIE Success Rate weighted by Path Length (RG SPL). We also report the performance of Navigation Success Rate, Navigation Oracle Success Rate, Navigation Success Rate weighted by Path Length (SPL), and Navigation Length. Please refer to the supplementary document for more details.

\subsection{Ablation Study}
\noindent In this section, we aim to answer the following questions: $(a)$ Does the performance gain mainly come from BERT-based structure? $(b)$ How effective is each of the proposed component? $(c)$ Does the memory blocks number matter? $(d)$ Why do we need logit fusion? For simplicity concern, we define the following experiment settings: $(1)$ our proposed $ViLEncoder$ is $ViL_{enc}$; $(2)$ the $ViLRaw_{enc}$ is $ViLEncoder$ not pre-trained on the Scene Grounding task but pre-trained on the Conceptual Captions dataset~\cite{sharma2018conceptual} as well as the $12$ tasks specified in ~\cite{lu2020multitask}; $(3)$ the $BERT_{enc}$ is a BERT language encoder pre-trained on the BookCorpus~\cite{zhu2015alignbooks} and English Wikipedia datasets; $(4)$ our proposed $ViLPointer$ is $ViL_{ptr}$; $(5)$ previous SOTA MattNet pointer is $MN_{ptr}$; $(6)$ our action policy is $MA_{pol}$; $(7)$ previous action policy is $C_{pol}$; $(8)$ previous simple language encoder is $L_{enc}$ composed of a trainable embedding layer with a Bi-directional LSTM layer.

\textbf{Performance Gain.} To answer question $(a)$, we perform experiments $1$, $2$, $3$ and $6$ as is shown in Table~\ref{tab:ablation}. All agents are trained under $\mathcal{L}_{ce}$ and $\mathcal{L}_{pm}$ with $\alpha$ and $\beta$ both set to $0.5$. It is clear that the agent's overall performance is incrementally improved by changing the language encoder from the simple $L_{enc}$ to our proposed $ViL_{enc}$, which proves our analysis that previous language encoder does not well capture the semantics of high-level instructions. The experimental results of $3$ and $6$ clearly suggests that the BERT-based structure is not the root cause of our performance gain and our proposed Scene Grounding task significantly increase the RG SPL metric to $33.9\%$ on Val Seen and $7.31\%$ on Val Unseen, even higher than the strong baseline in experiment $2$.

\textbf{Component Effectiveness.} To answer question $(b)$, based on the statistics from Table~\ref{tab:ablation}, we train six models in experiments from $3$ to $8$ and ablate the proposed component one by one to demonstrate the effectiveness. For fair comparison, we follow the settings of ~\cite{qi2020reverie}. All agents are trained under $\mathcal{L}_{ce}$ and $\mathcal{L}_{pm}$ with $\alpha$ and $\beta$ both set to $0.5$. We start from the baseline experiment $3$ and replace each component by our proposed ones. Specifically, in experiments $3$ and $6$, the proposed $ViLEncoder$ improves the RG SPL (and SPL) by a large margin, $11.98\%$ (and $21.52\%$) higher in Val Seen and $2.47\%$ (and $3.77\%$) higher in Val Unseen than the baseline respectively, which proves that the Scene Grounding task is effective; in experiments $3$ and $4$, our pointer $ViLPointer$ outperforms the MattNet counterpart by shortening the length of the search trajectory while maintaining a high RG SPL (and SPL), which demonstrates the effectiveness of the Object Grounding task; in experiments $3$ and $5$, the results show that the overall search trajectory of our action policy is longer than that of the baseline while our action policy achieves higher RGS and Navigation Success Rate, which demonstrates that the memory structure in our policy guides the agent to the correct target location at the cost of long trajectory; in experiments $7$ and $8$, we demonstrate that by integrating all our proposed methods, our agent improves previous SOTA in terms of RG SPL by $10.76\%$ on Val Seen and $5.6\%$ on Val Unseen.

\textbf{Memory Blocks.} To answer question $(c)$, we train five models with different $N_{mem}$ and $N_{state}$ values. In these experiments, we train the agents with $\mathcal{L}_{ce}$,  $\mathcal{L}_{pm}$ and $\mathcal{L}_{RL}$ and $\alpha$,  $\beta$ and $\gamma$ set to $1.0$. In general, according to the experiments from $9$ to $13$ in Table~\ref{tab:ablation}, all pairs of $(N_{mem}, N_{state})$ exhibit superior performance compared to previous SOTA method in experiment $1$ and the strong BERT baseline model in experiment $2$. Moreover, the best performance model is achieved by setting $(N_{mem}, N_{state})$ to $(3, 3)$ in these five models, which suggests that using small values of $(N_{mem}, N_{state})$ limits the agent's memorization ability and using large values of $(N_{mem}, N_{state})$ enables the agent to achieve good performance on Val Unseen while maintains good performance on Val Seen.


\begin{table}[t]\small
\centering
\caption{Pointer Task: REVERIE Success Rate at the ground truth target viewpoint; Encoder Task: given ground truth path, the success rate of identifying the target viewpoint among a set of candidate viewpoints along the path.}
\resizebox{\linewidth}{!}{
\begin{tabular}{c|ccc}
\hline
Tasks & Methods & Val Seen & Val Unseen \\
\hline
\multirow{4}{*}{Pointer} & MattNet~\cite{yu2018mattnet} & 68.45 & 56.63 \\
 & CM-Erase~\cite{liu2019improving} & 65.21 & 54.02 \\
 & ViLPointer-image-based & 65.72 & 55.53 \\
 & ViLPointer-vp-based & \underline{\textbf{73.26}} & \underline{\textbf{67.45}} \\
\hline
Encoder & ViLEncoder & \underline{\textbf{85.67}} & \underline{\textbf{66.43}} \\
\hline
\end{tabular}}
\label{tab:refer_and_encoder}
\end{table}


\textbf{Logit Fusion.} To answer question $(d)$, we report two accuracies to verify the effectiveness of $g_{og}$ and $g_{sg}$. In the Encoder Task of Table~\ref{tab:refer_and_encoder}, given ground-truth path, our proposed $ViLBERT$ model achieves competitive performance on both Val Seen and Val Unseen, demonstrating the strong ability of $g_{sg}$ to identify a target viewpoint. In the Pointer Task of Table~\ref{tab:refer_and_encoder}, the performance of $ViLPointer$-vp-based is significantly higher than previous image-based pointers because it is able to capture cross-image objects relationships, suggesting that $g_{og}$ has the ability to find the target location if the target object exists. According to experiments from $14$ to $17$, where the agents are trained with $\mathcal{L}_{ce}$,  $\mathcal{L}_{pm}$ and $\mathcal{L}_{RL}$ and $\alpha$,  $\beta$ and $\gamma$ set to $1.0$, summing $\boldsymbol{l}_{\tau}$, $g_{og}^{\tau}$, and $g_{sg}^{\tau}$ shortens the search trajectory and maintains a high RGS(Navigation Success Rate) and RG SPL(SPL). The motivation behind the summing strategy is to use model ensemble to reduce bias when searching for target locations considering the fact that the agent has no prior knowledge of the surrounding environments and the guidance of the high-level instructions is weak.

\begin{figure}[ht]
  \centering
  \includegraphics[width=\linewidth]{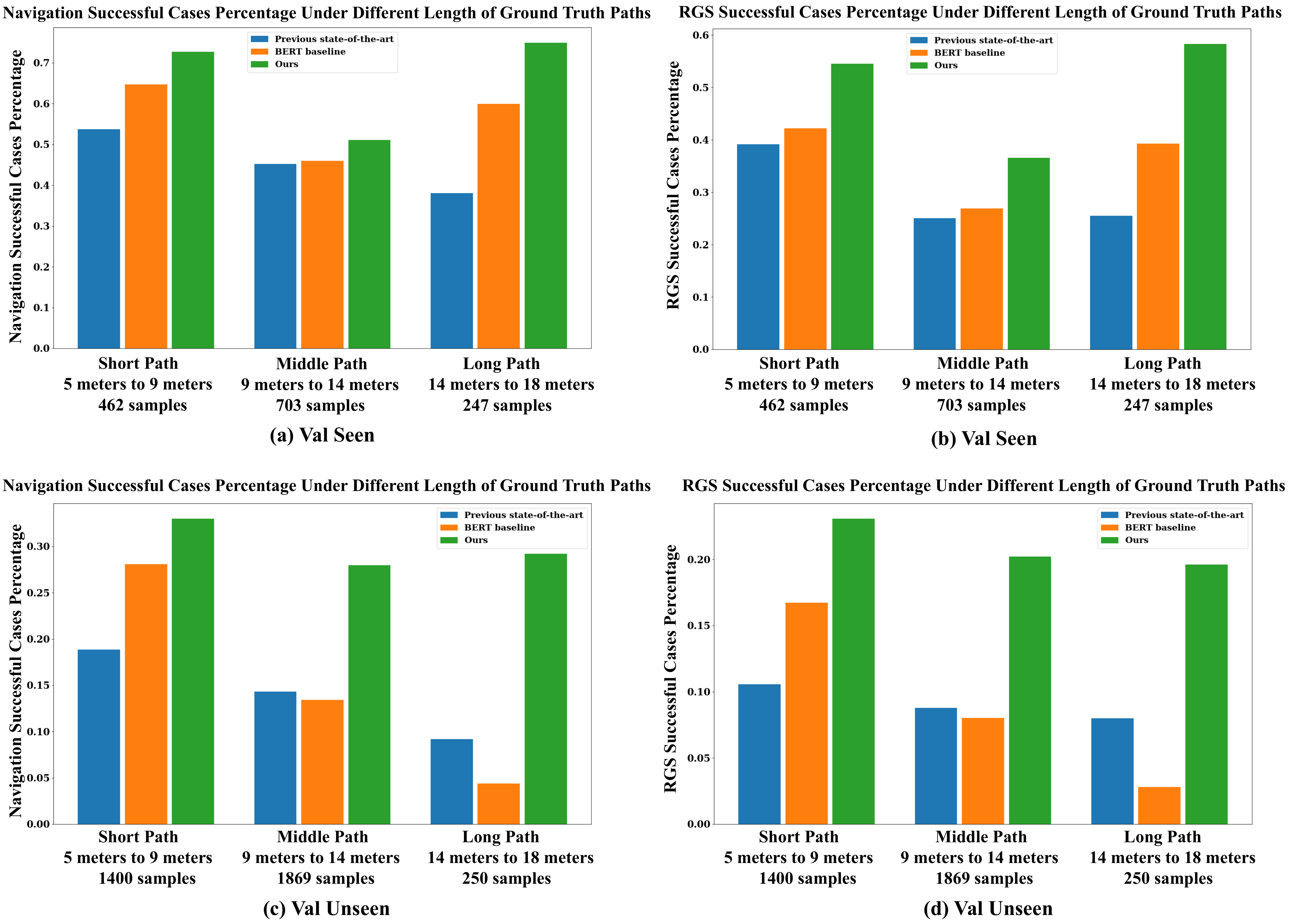}
  \caption{Percentage of successful Navigation and RGS cases under different length of ground-truth paths on Val Seen and Val Unseen datasets for \textcolor{oceanboatblue}{previous state-of-the-art method}, \textcolor{orange}{BERT baseline in experiment $2$}, and \textcolor{ForestGreen}{our method}.}
  \label{Fig:stats}
\end{figure}

\subsection{Compared to previous state-of-the-art results}
\noindent We first show what kind of cases our method improves compared to previous SOTA and our BERT-based strong baseline in experiment $2$. Specifically, we divide the shortest distance lengths of all ground-truth paths into three groups, namely short path($5$ meters to $9$ meters with $462$ sample paths on Val Seen and $1400$ sample paths on Val Unseen), middle path($9$ meters to $14$ meters with $703$ sample paths on Val Seen and $1869$ sample paths on Val Unseen), and long path($14$ meters to $18$ meters with $247$ sample paths on Val Seen and $250$ sample paths on Val Unseen). Then, we count the cases that the agent successfully navigates to the target locations and the cases that the agent successfully navigates and localizes the target objects for the three groups. In Fig.~\ref{Fig:stats}, we report the corresponding successful cases percentage. It is obvious that our proposed method improves all kinds of sample paths by a clear margin.

Then, we compare our final model with previous SOTA models in Table~\ref{tab:sota}. As is clearly shown in Table~\ref{tab:sota}, our model outperforms all previous models by a large margin. Specifically, in terms of SPL, our agent increases previous SOTA by $11.58\%$ on Val Seen, $9.09\%$ on Val Unseen and $3.24\%$ on Test respectively; for RG SPL, our agent increase previous SOTA by $12.99\%$ on Val Seen, $6.89\%$ on Val Unseen and $3.12\%$ on Test. The overall improvements indicate that our proposed scene-intuitive agent not only navigates better but also localizes target objects more accurately.

\section{Conclusion}
\noindent In this paper, we present a scene-intuitive agent capable of understanding high-level instructions for the REVERIE task. Different from previous works, we propose two pre-training tasks, Scene Grounding task and Object Grounding task respectively, to help the agent learn \textbf{\emph{where}} to navigate and \textbf{\emph{what}} object to localize simultaneously. Moreover, the agent is trained with a Memory-augmented action decoder that fuses grounded textual representation and visual representation with memory augmented current state representation to generate action sequence. We extensively verify the effectiveness of our proposed components and experimental results demonstrate that our result outperforms previous methods significantly. Nevertheless, how to bridge the performance gap between seen and unseen environments and how to shorten the navigation length efficiently remains an open problem for further investigation.

{\small
\bibliographystyle{ieee_fullname}
\bibliography{egbib}
}

\clearpage

\section{More Related Work}
\noindent\textbf{Behavioral Research on Human Navigation.} The behavioural research of navigation of human beings has a long history and is still under active research~\cite{Ham2020LargescaleAO, Jetzschke2017FindingHL, Wolbers2010WhatDO, Wiener2009TaxonomyOH, Chrastil2013NeuralES, Coutrot2018GlobalDO, Gopal1989NavigatorAP}. Yet, it is not well understood how we human carry out the learning process of navigation in our brain to allow us to navigate in a familiar or unfamiliar environment. However, according to ~\cite{Ham2020LargescaleAO, Wolbers2010WhatDO, Wiener2009TaxonomyOH, Chrastil2013NeuralES}, we humans use a range of different cognitive processes when we navigate. For example, we identify representative landmark cues, memorize our goal location, and identify the shortest route to that goal location. A significant number of research have supported such dissociable cognitive aspects. The human intuitions for remote embodied navigation we referred to in this paper is a set of commonsense rules and heuristics that come from observations of humans¡¯ life experiences, which shares a similar motivation mentioned in ~\cite{Gupta2019CognitiveMA}. Our work has also proved that drawing on such observations in high-level VLN is a promising direction.

\section{Implementation Details}
In this section, we introduce the implementation details of the pre-training stage and the action decoding stage. In pre-training stage, we first present the sampled datasets information of the Scene Grounding task and the Object Grounding task. Second, we introduce the ViLBERT model used in the pre-training stage. Third, we illustrate the action decoder architecture and the training parameters in detail.

\subsection{Pre-training Stage Details}
\noindent\textbf{Scene Grounding Task.} The Scene Grounding training dataset consists of $10312$ samples, each containing an instruction and four viewpoints out of which one is positive. The sampling strategy is illustrated in the main paper. We evaluate the effectiveness of this task by asking the model trained to identify the true target viewpoint given the ground-truth path. We report the accuracy on the Val Seen ($1423$ paths) and Val UnSeen ($3521$ paths) REVERIE.

\noindent\textbf{Object Grounding Task.} The image based grounding dataset contains $67432$ training samples and the viewpoint based object grounding dataset contains $4356$ training samples. The sampling strategy is presented in the main paper. Similar to ~\cite{qi2020reverie}, we evaluate the performance of this model on the ground-truth target viewpoint and report the object grounding accuracy.

\noindent\textbf{Model Details.} The ViLBERT model used in Scene Grounding task and Object Grounding task consists of a language stream, a vision stream and a cross modal alignment layers block. The language stream utilizes a $BERT_{BASE}$ architecture~\cite{devlin2018bert}, which has $12$-layer of transformer blocks and each block having a hidden state size of $768$ and $12$ attention heads. The vision stream and the cross modal alignment block use $6$-layer transformer blocks and each having a hidden state size of $1024$ and $8$ attention heads respectively. Following ~\cite{lu2019vilbert, lu2020multitask}, the language stream is initialized with BERT weights pre-trained on the BookCorpus~\cite{zhu2015alignbooks} and English Wikipedia datasets. Then, the ViLBERT model is pre-trained on the Conceptual Captions dataset~\cite{sharma2018conceptual} as well as the $12$ tasks specified in ~\cite{lu2020multitask}. Finally, it is fine-tuned on our Scene Grounding task and Object Grounding task respectively. In the Scene Grounding task, the Scene Grounding model is trained with the Adam optimizer with a learning rate of $4e-5$ and a batch size of $32$ for $10$ epochs. In the Object Grounding task, the Object Grounding model is first trained on the image based Object Grounding dataset with the Adam optimizer with a learning rate of $4e-5$ and a batch size of $128$ for $20$ epochs. Then, it is further fine-tuned on the viewpoint based Object Grounding dataset with the Adam optimizer with a learning rate of $1e-5$ and a batch size of $128$ for $10$ epochs. We use a linear decay learning rate schedule with warm up to train the aforementioned models. All models are trained on NVIDIA Geforce $2080$Ti GPUs with $11$GB memory using Pytorch ~\cite{NEURIPS2019_9015}.

\subsection{Action Decoding Stage Details}
\noindent The $ViLEncoder$ is composed of a ViLBERT model pre-trained on the Scene Grounding task and a Bi-directional LSTM layer. The $D_{h}$ in the BiLSTM is set to $512$. The $ViLPointer$ is ViLBERT model pre-trained on the Object Grounding task. The $N_{mem}$ and $N_{state}$ used in the memory blocks are set to $3$ according to our ablation study in the ablation study. We follow the same RL setting as ~\cite{tan2019envdrop} that sets the discounted factor to $0.9$ and adopts reward shaping ~\cite{wu2018building}. We train the agent with the Adam Optimizer ~\cite{dk2015adam} with a learning rate of $1e-4$, weight decay of $5e-4$, batch size of $64$ and the maximum decoding action length of $40$. We clip the global gradient norm at $40$. We train the agent for $13000$ iterations and report the final performance. All experiments have been conducted on NVIDIA Geforce $2080$Ti GPUs with $11$GB memory using Pytorch ~\cite{NEURIPS2019_9015}.

\section{Evaluation Metrics Details}
In this section, we illustrate the details of the evaluation metrics. Following ~\cite{qi2020reverie}, we evaluate the performance of the model based on REVERIE Success Rate (RGS) and REVERIE Success Rate weighted by Path Length (RG SPL). Besides, we report the performance of our method on the following metrics in the REVERIE dataset. It is worth noting that the target object is only observable within $3$ meters of the target viewpoint.
\begin{itemize}
    \item \noindent\textbf{Navigation Success Rate} is the percentage of the target object observable at the agent's final location.
    \item \noindent\textbf{Navigation Oracle Success Rate} measures the percentage of the target object that can be observed at one of the agent's passed viewpoints.
    \item \noindent\textbf{Navigation Success Rate weighted by Path Length (SPL)} is the navigation success rate weighted by the trajectory length.
    \item \noindent\textbf{Navigation Length} is the trajectory length in meters.
    \item \noindent\textbf{REVERIE Success Rate (RGS)} is calculated as the percentage of the output bounding box that has an IoU $\geq 0.5$ with the ground truth box.
    \item \noindent\textbf{REVERIE Success Rate weighted by Path Length (RG SPL)} is REVERIE success rate weighted by the trajectory length.
\end{itemize}

\section{Qualitative Examples}
\begin{figure*}[ht]
  \centering
  \includegraphics[width=0.75\linewidth]{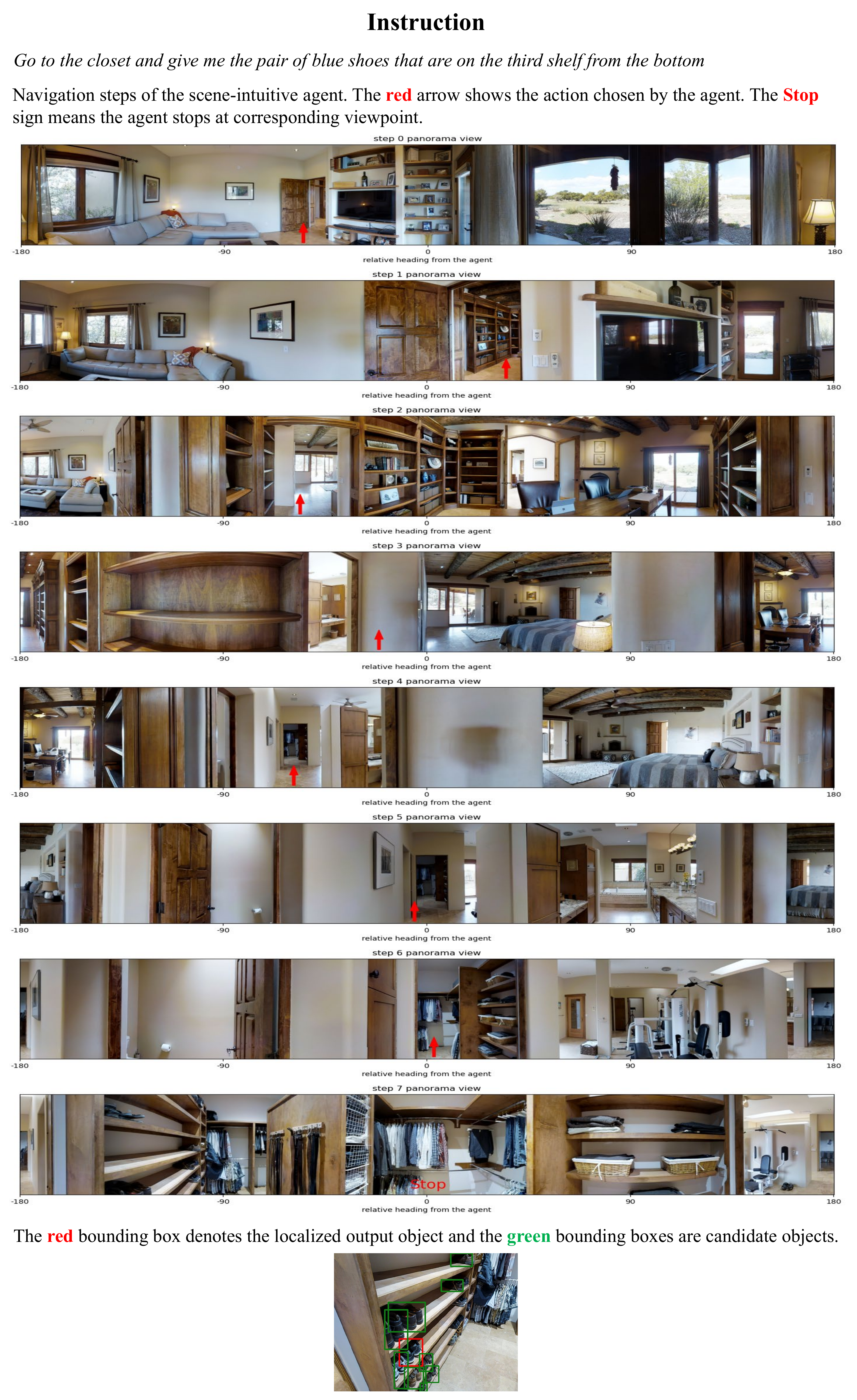}
  \caption{The successful navigation and localization qualitative example result on Val Seen dataset.}
  \label{Fig:val_seen_fig_1}
\end{figure*}

\begin{figure*}[ht]
  \centering
  \includegraphics[width=0.75\linewidth]{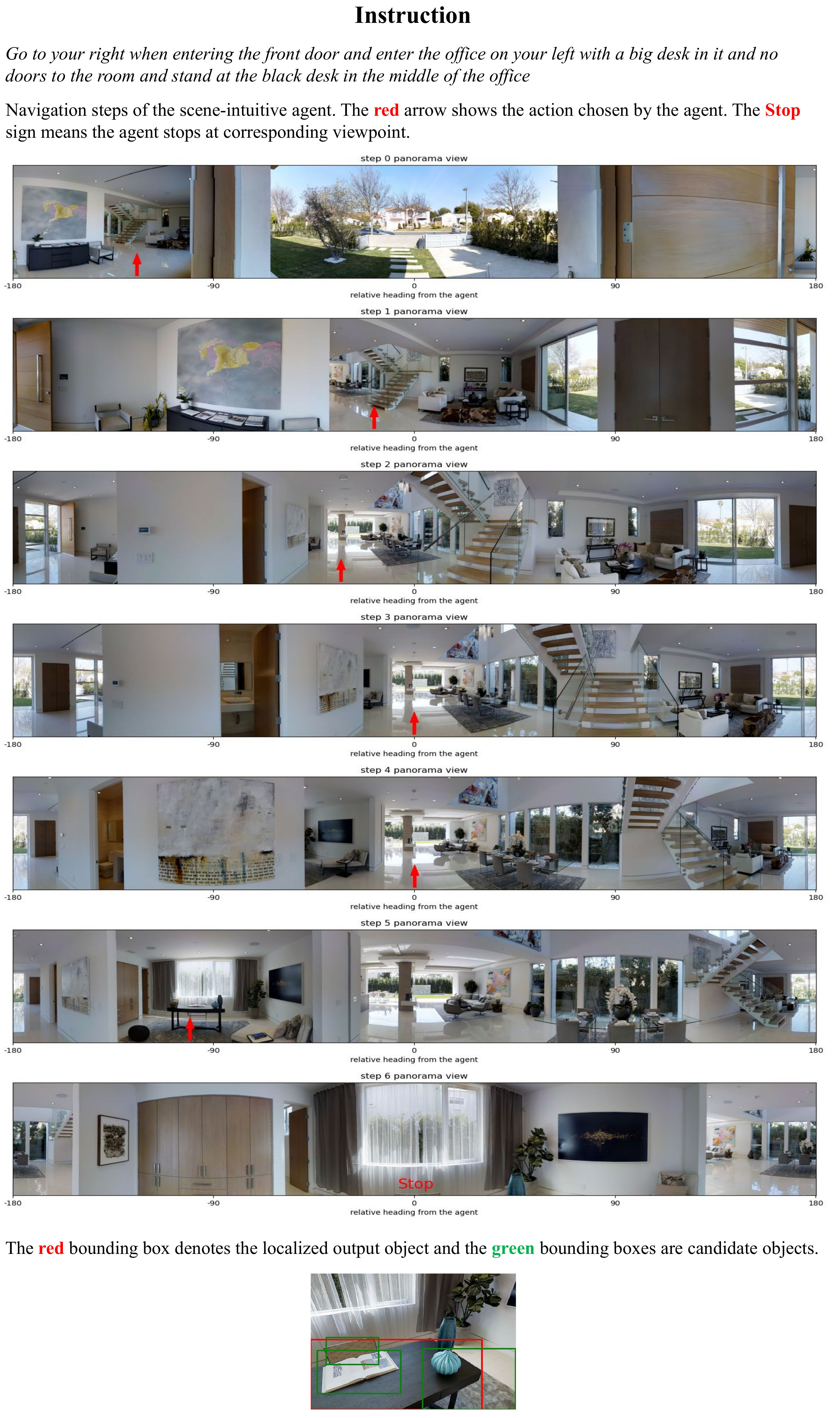}
  \caption{The successful navigation and localization qualitative example result on Val Seen dataset.}
  \label{Fig:val_seen_fig_2}
\end{figure*}

\begin{figure*}[ht]
  \centering
  \includegraphics[width=0.75\linewidth]{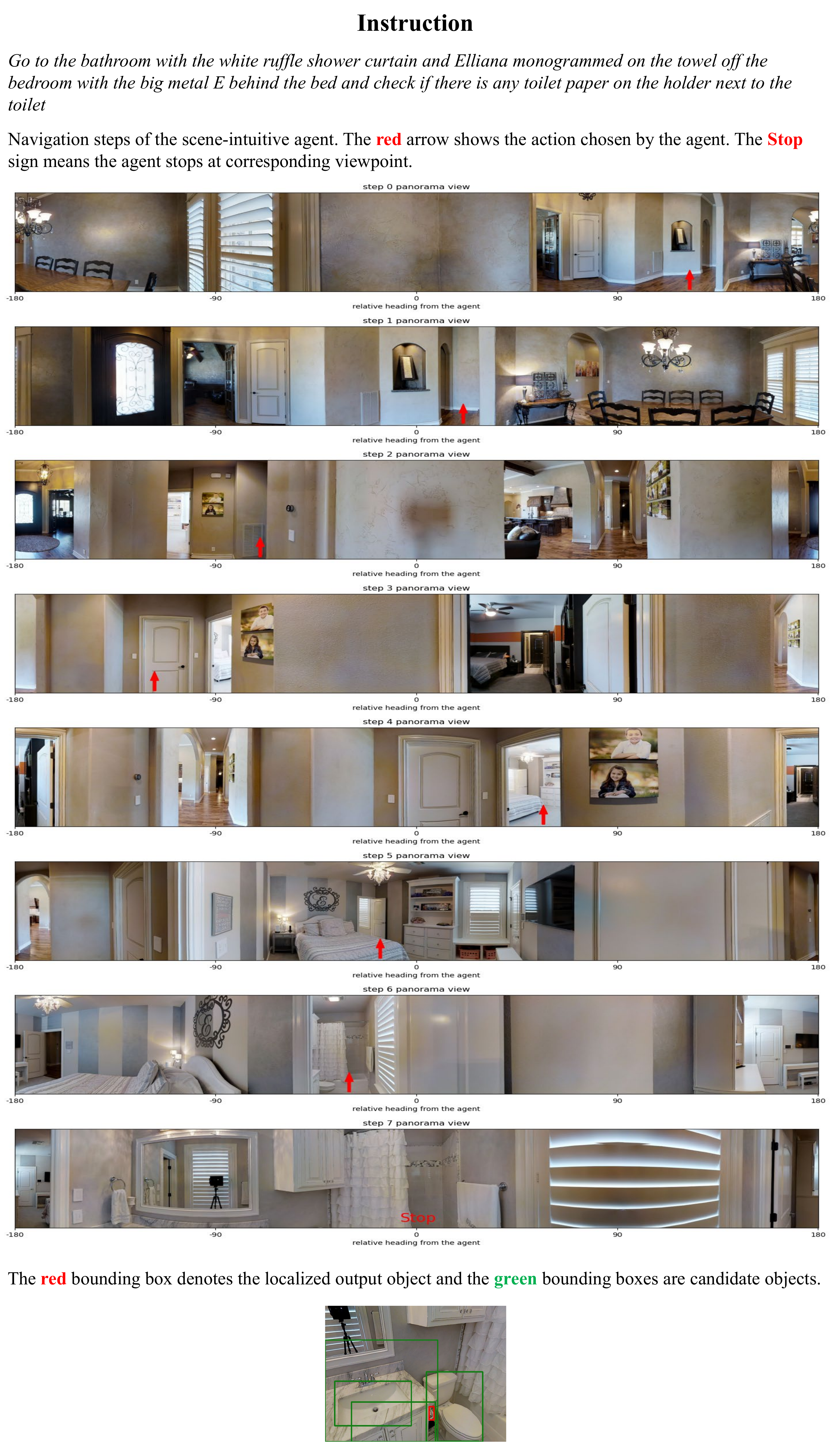}
  \caption{The successful navigation and localization qualitative example result on Val Seen dataset.}
  \label{Fig:val_seen_fig_3}
\end{figure*}

\begin{figure*}[ht]
  \centering
  \includegraphics[width=0.75\linewidth]{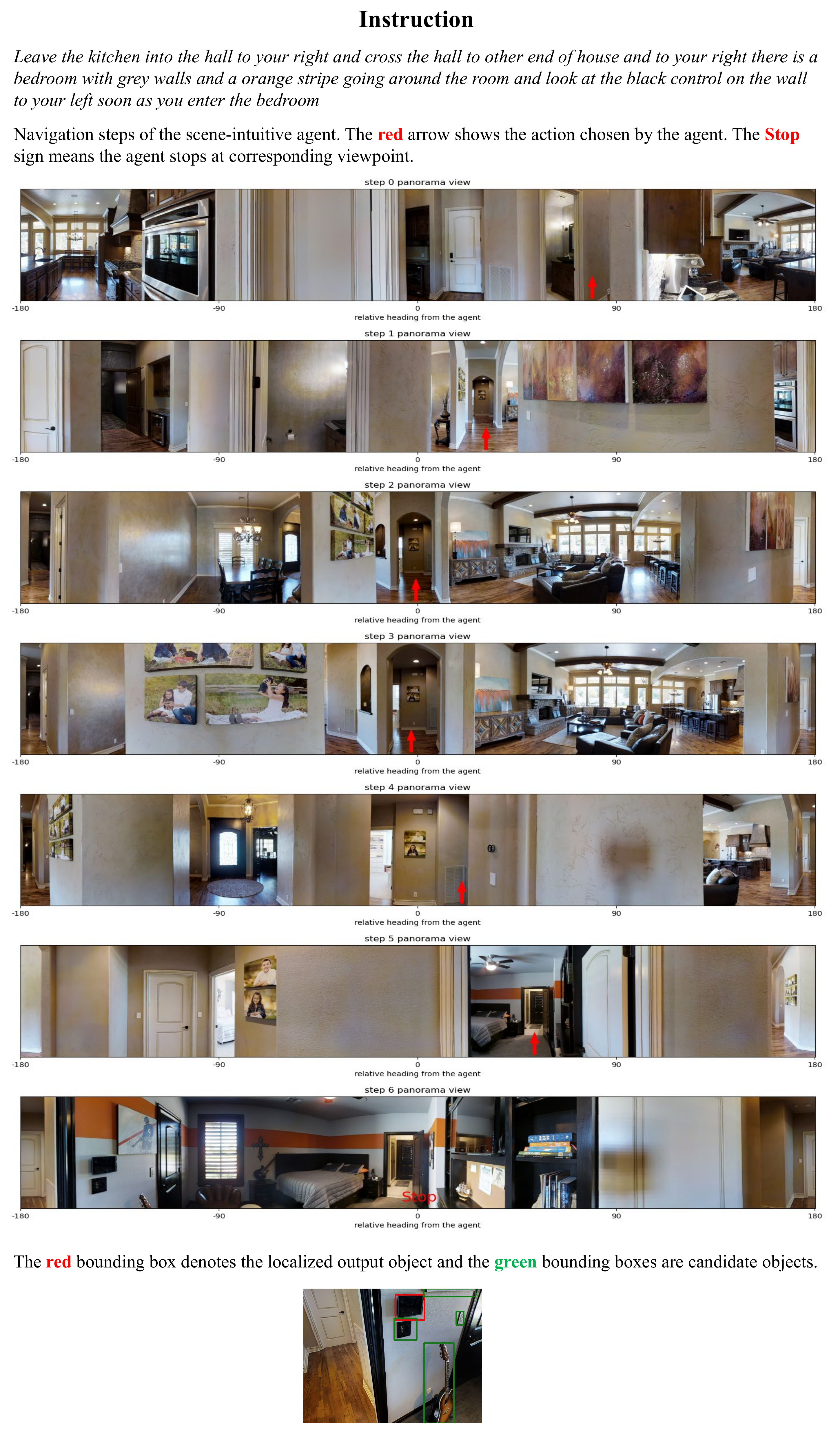}
  \caption{The successful navigation and localization qualitative example result on Val Seen dataset.}
  \label{Fig:val_seen_fig_4}
\end{figure*}

\begin{figure*}[ht]
  \centering
  \includegraphics[width=0.75\linewidth]{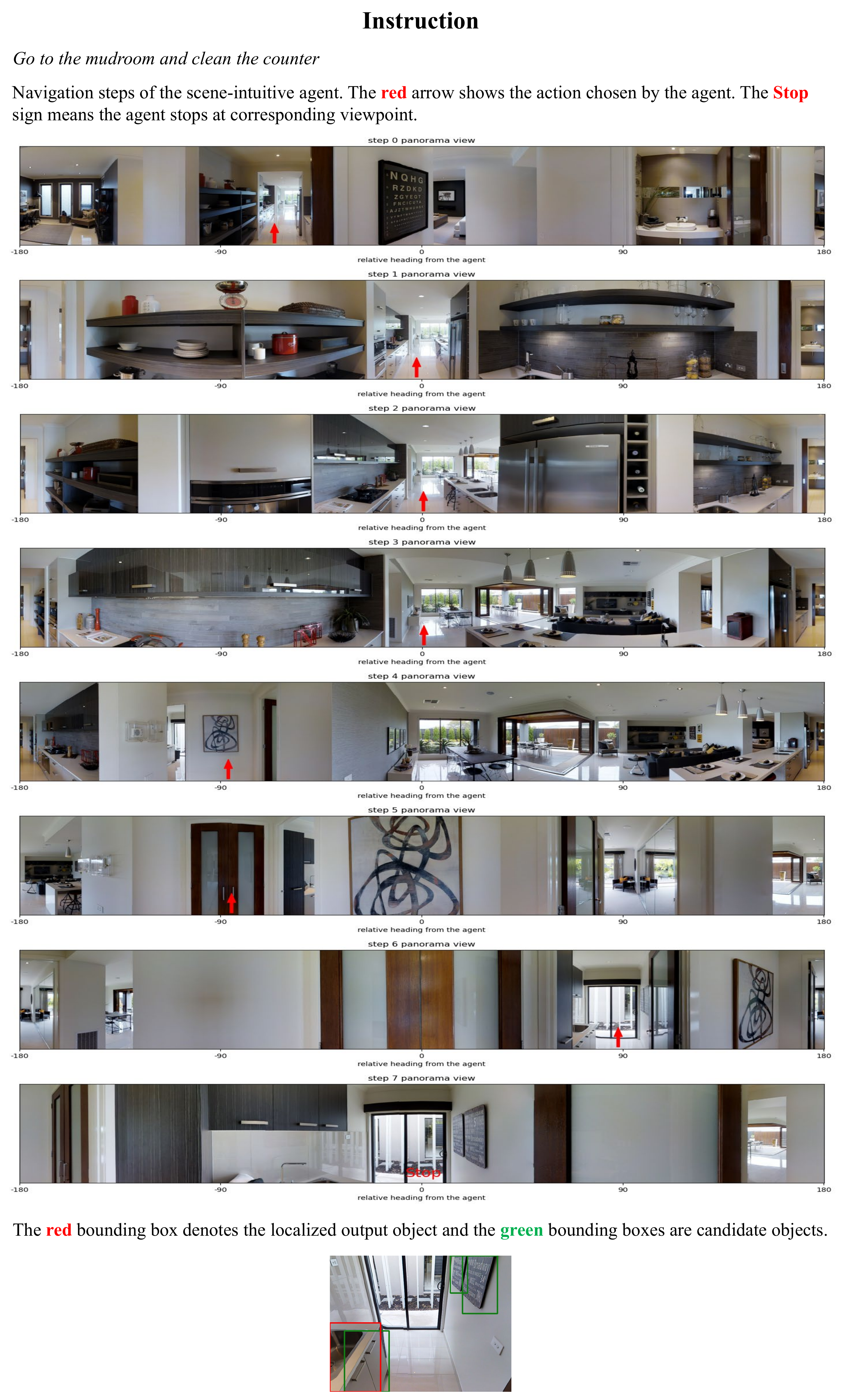}
  \caption{The successful navigation and localization qualitative example result on Val Unseen dataset.}
  \label{Fig:val_unseen_fig_1}
\end{figure*}

\begin{figure*}[ht]
  \centering
  \includegraphics[width=0.75\linewidth]{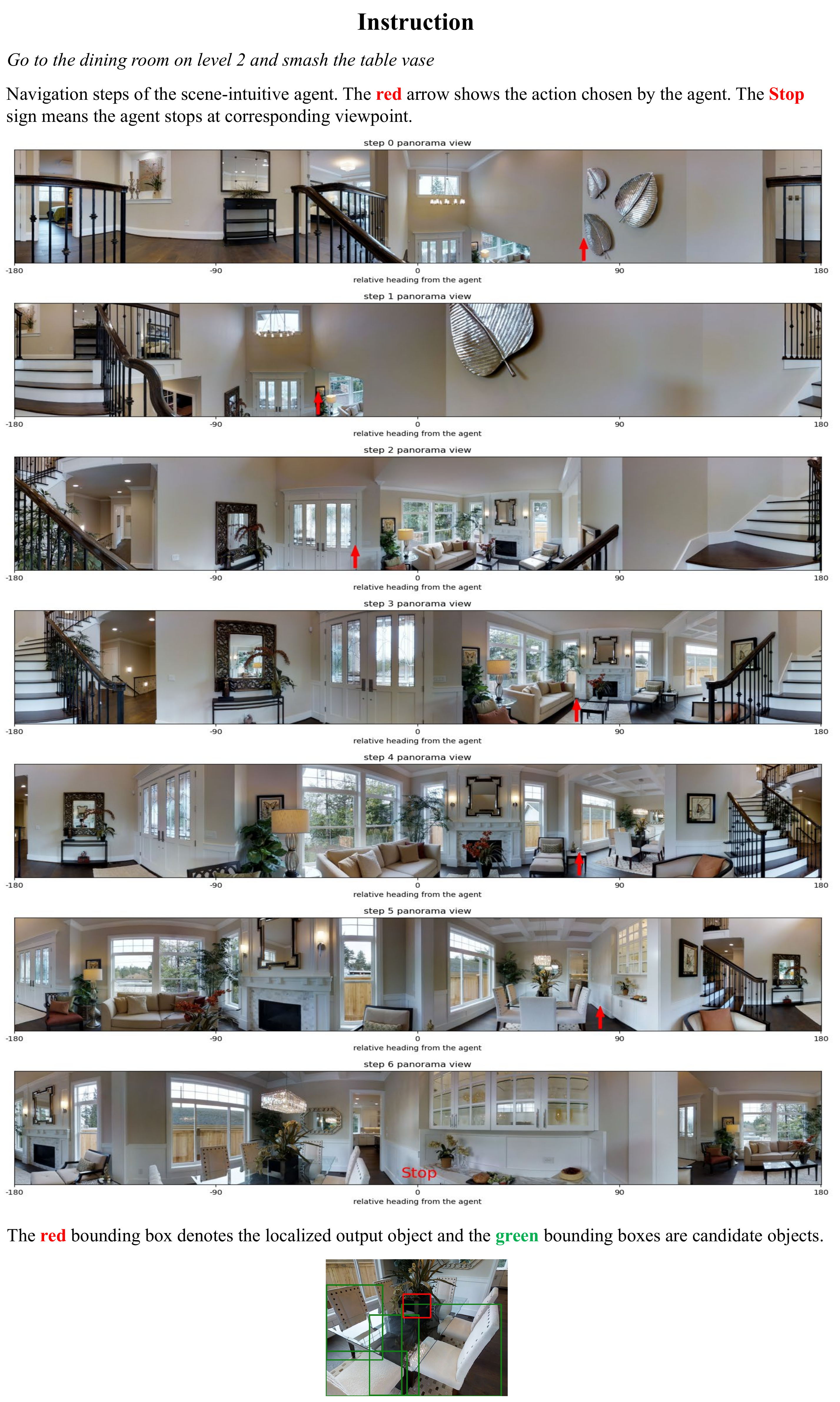}
  \caption{The successful navigation and localization qualitative example result on Val Unseen dataset.}
  \label{Fig:val_unseen_fig_2}
\end{figure*}

\begin{figure*}[ht]
  \centering
  \includegraphics[width=0.75\linewidth]{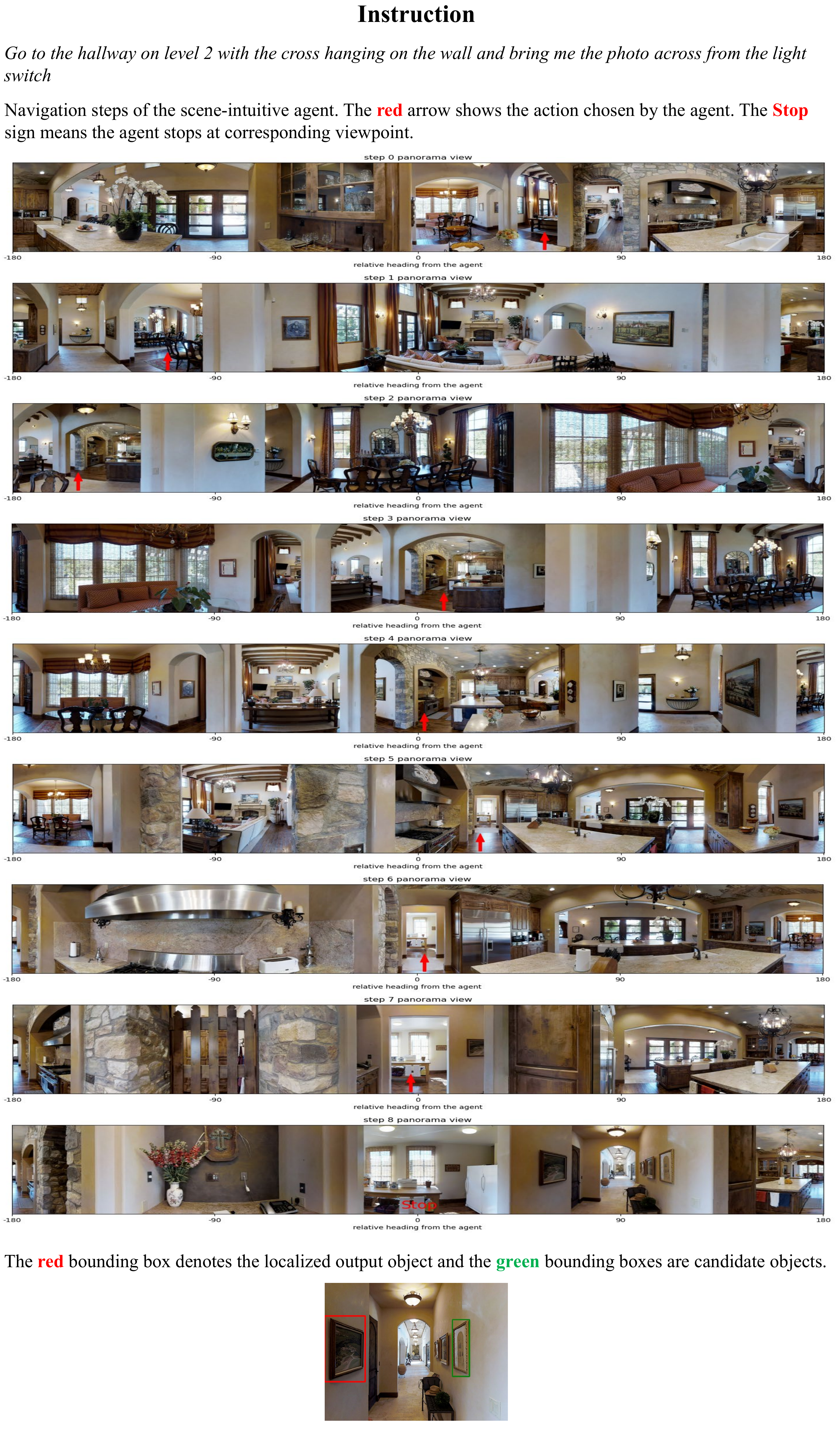}
  \caption{The successful navigation and localization qualitative example result on Val Unseen dataset.}
  \label{Fig:val_unseen_fig_3}
\end{figure*}

\begin{figure*}[ht]
  \centering
  \includegraphics[width=0.75\linewidth]{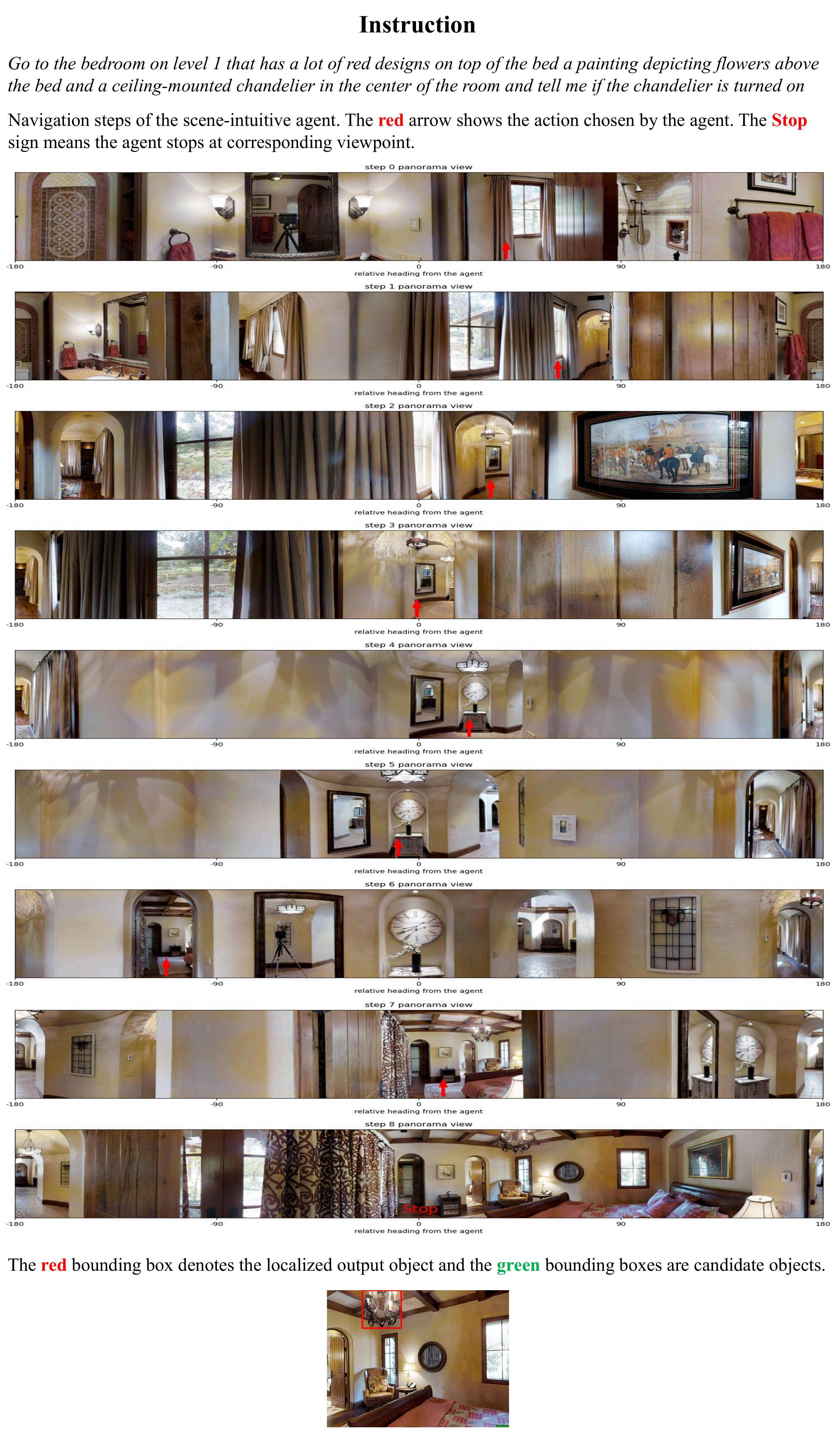}
  \caption{The successful navigation and localization qualitative example result on Val Unseen dataset.}
  \label{Fig:val_unseen_fig_4}
\end{figure*}

In this section, we show a number of qualitative examples of how our proposed agent performs in both Val Seen environment (from Fig.~\ref{Fig:val_seen_fig_1} to Fig.~\ref{Fig:val_seen_fig_4}) and Val Unseen environment (from Fig.~\ref{Fig:val_unseen_fig_1} to Fig.~\ref{Fig:val_unseen_fig_4}). Besides, we also visualize five representative failed cases illustrating the typical mistakes our agent make to better understand how our agent works.

\begin{figure*}[ht]
  \centering
  \includegraphics[width=0.95\linewidth]{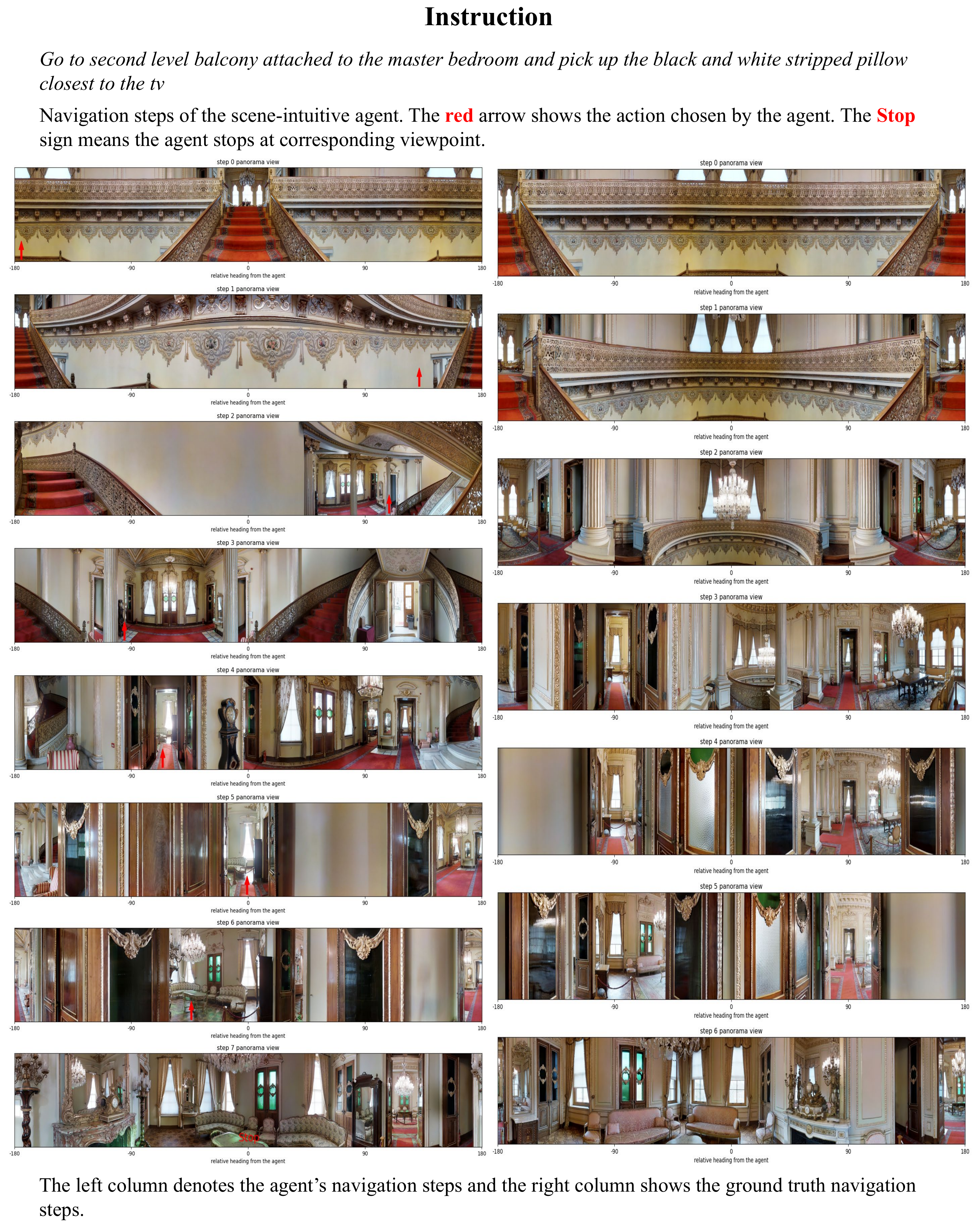}
  \caption{The failed navigation qualitative example result. In this example, the agent first successfully navigates to second level but failed to enter the correct bedroom and stopped at a wrong viewpoint.}
  \label{Fig:failed_fig_1}
\end{figure*}

\begin{figure*}[ht]
  \centering
  \includegraphics[width=0.75\linewidth]{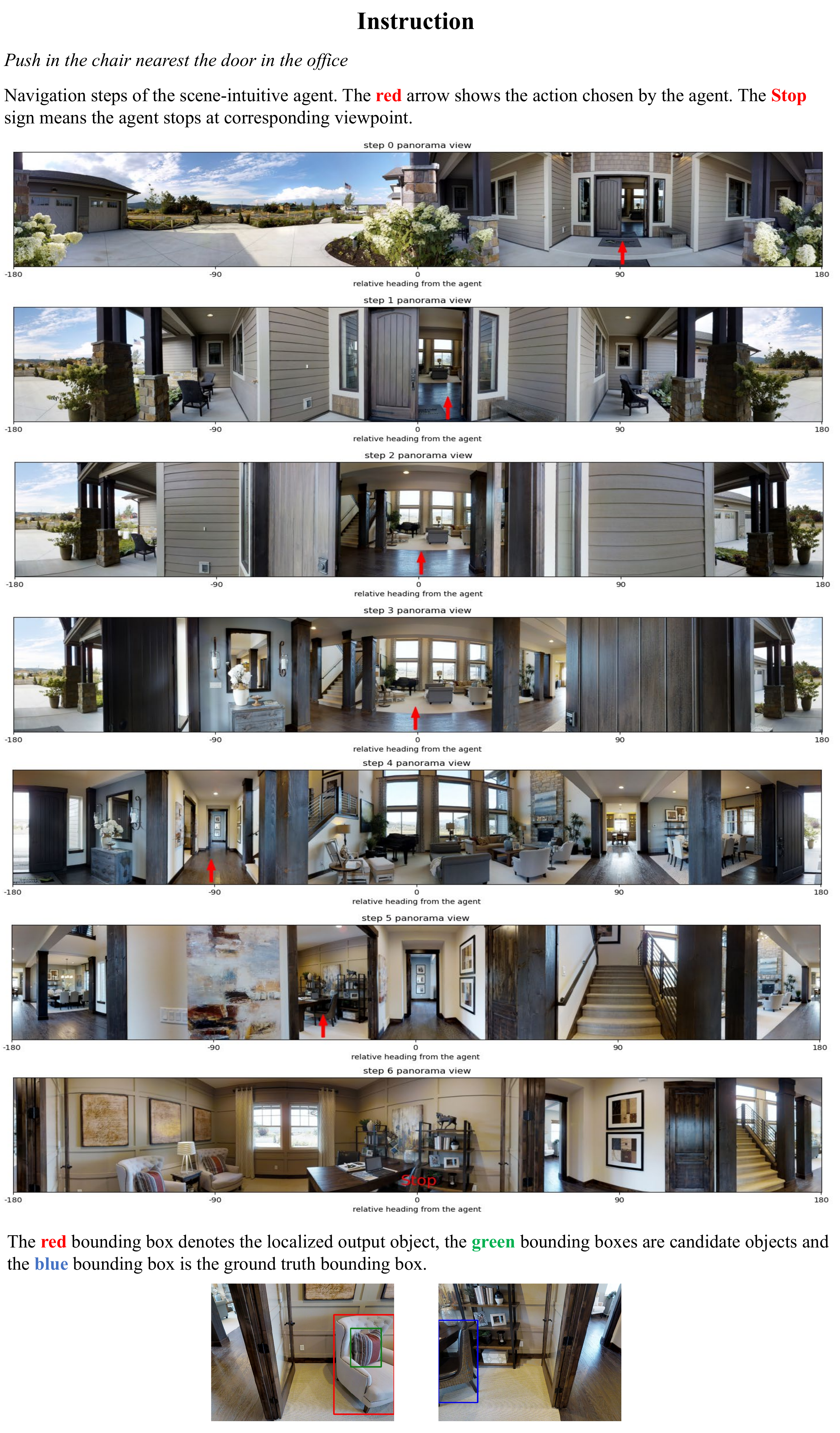}
  \caption{The failed localization qualitative example result. In this example, the agent first successfully navigates to the target viewpoint but failed to localize the target object because the $ViLPointer$ module thinks the white chair is closer to the office door than the black chair, which is reasonable as it is hard to decide which one is closer.}
  \label{Fig:failed_fig_1}
\end{figure*}

\begin{figure*}[ht]
  \centering
  \includegraphics[width=0.75\linewidth]{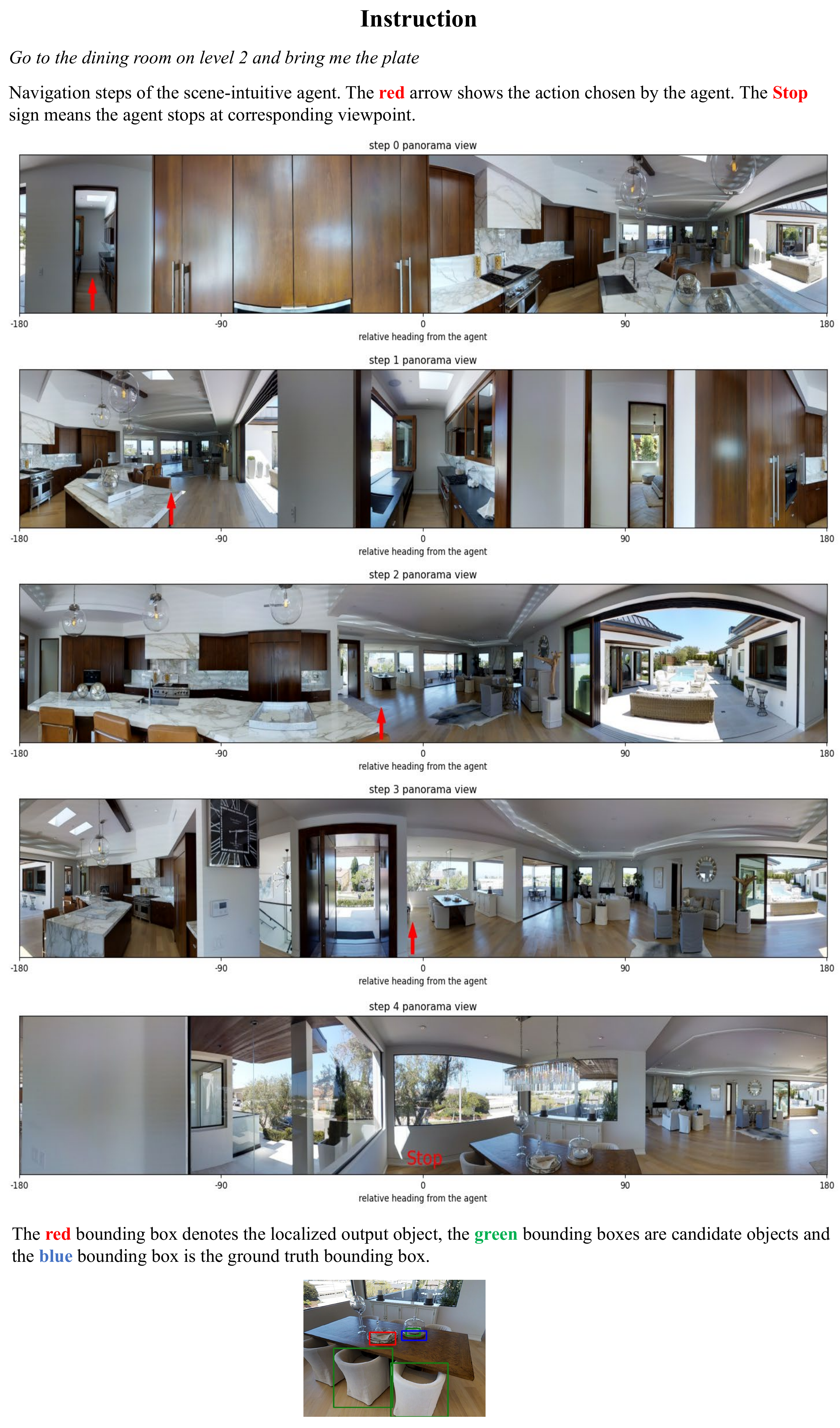}
  \caption{The failed localization qualitative example result. In this example, the agent first successfully navigates to the target viewpoint but failed to localize the target object because of the ambiguous meaning of ``the plate'' in the high-level instruction.}
  \label{Fig:failed_fig_1}
\end{figure*}

\begin{figure*}[ht]
  \centering
  \includegraphics[width=0.75\linewidth]{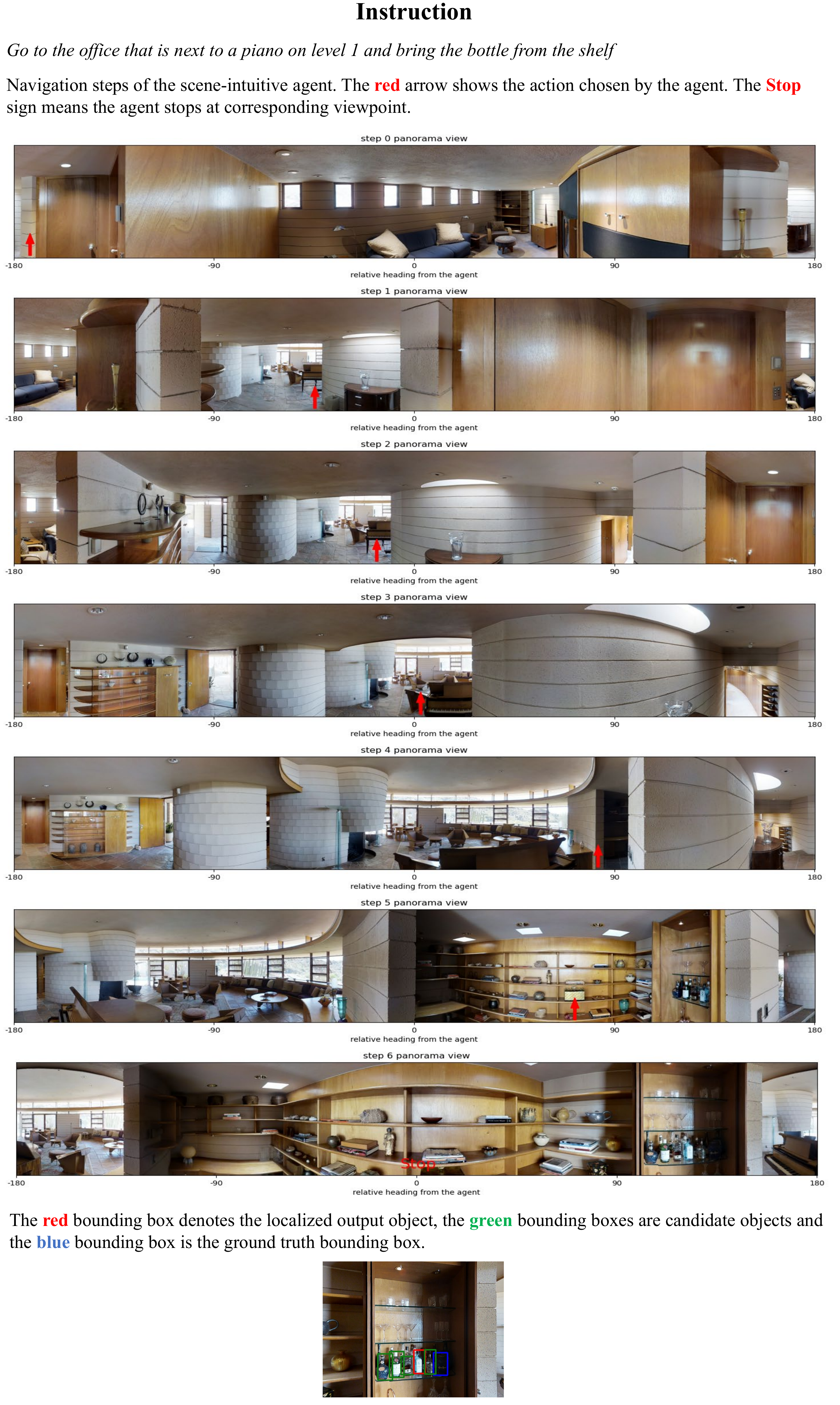}
  \caption{The failed localization qualitative example result. In this example, the agent first successfully navigates to the target viewpoint but failed to localize the target object because of the ambiguous meaning of ``the bottle on the shelf'' in the high-level instruction.}
  \label{Fig:failed_fig_1}
\end{figure*}

\begin{figure*}[ht]
  \centering
  \includegraphics[width=0.75\linewidth]{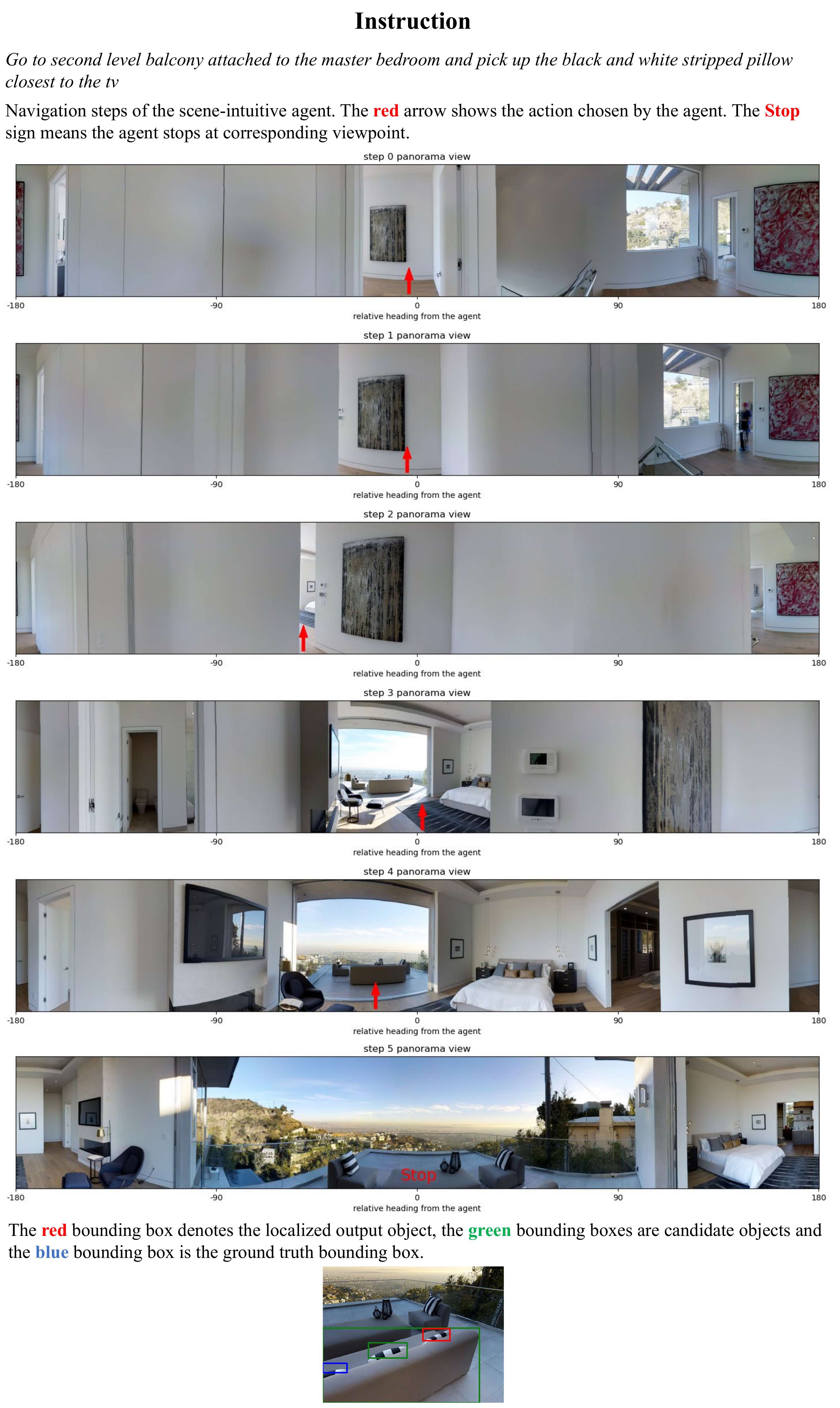}
  \caption{The failed localization qualitative example result. In this example, the agent first successfully navigates to the target viewpoint but failed to localize the target object because it failed to capture the relative position of similar objects (``the black and white stripped pillow'') in the scene.}
  \label{Fig:failed_fig_1}
\end{figure*}

\end{document}